\documentclass{article}

\usepackage{PRIMEarxiv}

\usepackage{algorithm}
\usepackage{algpseudocode}

\usepackage[english]{babel}
\addto\captionsenglish{
	
}

\usepackage{hyperref}       
\usepackage{url}            
\usepackage{amsfonts}  
\usepackage{enumerate}
\usepackage{verbatim}
\usepackage{amsmath,amsthm}
\usepackage{multirow}
\usepackage{epsfig}
\usepackage{booktabs}
\usepackage{graphicx} 
\usepackage{subcaption}
\usepackage{makecell}
\usepackage{colortbl}  
\usepackage{xcolor}
\usepackage{array}   
\usepackage{caption}
\usepackage{cleveref}
\crefname{figure}{Fig.}{Figs.}
\Crefname{figure}{Fig.}{Figs.}
\crefname{table}{Table}{Tables}
\Crefname{table}{Table}{Tables}

\usepackage{titlesec}
\usepackage{bbding}

\usepackage{makecell}
\usepackage{rotating}

\titleformat{\paragraph}
{\normalfont\normalsize\bfseries}{\theparagraph}{0.5em}{}
\titlespacing*{\paragraph}
{0pt}{3.25ex plus 1ex minus .2ex}{1.5ex plus .2ex}

\usepackage[utf8]{inputenc} 
\usepackage[T1]{fontenc}    
\usepackage{hyperref}       
\usepackage{url}            
\usepackage{booktabs}       
\usepackage{amsfonts}       
\usepackage{nicefrac}       
\usepackage{microtype}      
\usepackage{lipsum}
\usepackage{fancyhdr}       
\usepackage{graphicx}       
\graphicspath{{media/}}     

\title{Aligning First, Then Fusing: A Novel Weakly Supervised Multimodal Violence Detection Method
}

\author{
  Wenping Jin, Li Zhu \\
  School of Software Engineering \\
  Xi'an Jiaotong University \\
  Xi'an, China\\
  \texttt{jinwenping@stu.xjtu.edu.cn, zhuli@xjtu.edu.cn} \\
   \And
  Jing Sun \\
  Shenzhen Institute of Advanced Technology \\
  Chinese Academy of Sciences \\
  Shenzhen, China\\
  \texttt{jing.sun1@siat.ac.cn} \\
}

\begin{document}
\maketitle

\begin{abstract}
Weakly supervised violence detection refers to the technique of training models to identify violent segments in videos using only video-level labels. Among these approaches, multimodal violence detection, which integrates modalities such as audio and optical flow, holds great potential. Existing methods in this domain primarily focus on designing multimodal fusion models to address modality discrepancies. In contrast, we take a different approach—leveraging the inherent discrepancies across modalities in violence event representation to propose a novel multimodal semantic feature alignment method. This method sparsely maps the semantic features of local, transient, and less informative modalities (such as audio and optical flow) into the more informative RGB semantic feature space. Through an iterative process, the method identifies the suitable‌ non-zero feature matching subspace and aligns the modality-specific event representations based on this subspace, enabling the full exploitation of information from all modalities during the subsequent modality fusion stage. Building on this, we design a new weakly supervised violence detection framework that consists of unimodal multiple-instance learning for extracting unimodal semantic features, multimodal alignment, multimodal fusion, and final detection. Experimental results on benchmark datasets demonstrate the effectiveness of our method, achieving an average precision (AP) of 86.07\% on the XD-Violence dataset. Our code is available at https://github.com/xjpp2016/MAVD.
\end{abstract}


\section{Introduction}
\label{s1}
Violence Detection (VD) aims to identify violent events in videos, offering significant potential for application in fields such as security surveillance and content moderation\cite{2011_VD,fast_VD}. However, in the supervised learning paradigm, accurately locating violent events requires frame-by-frame annotation, which is both time-consuming and labor-intensive. To overcome this challenge, many recent studies have adopted weakly supervised learning frameworks based on Multi-Instance Learning (MIL)\cite{1997_MIL}. MIL-based VD methods treat videos as bags, with video-level labels indicating the presence or absence of violence, and learn to identify the top-K most discriminative instances within each bag.
\par
Currently, most weakly supervised VD methods primarily focus on visual tasks \cite{MIL_VVD_1,MIL_VVD_2,MIL_VVD_3,MIL_VVD_4,MIL_VVD_5}, with relatively limited research on multimodal approaches that incorporate audio \cite{xd_violence,xd_violence_TMM}. However, multimodal VD holds significant potential, as audio offers valuable complementary insights. In particular, certain sounds—such as shouting, fighting noises, gunshots, or explosions—frequently accompany violent events.
\par
The core challenge of weakly supervised multimodal VD lies in effectively integrating information from different modalities in the absence of detailed label information, which can also be framed as the challenge of achieving effective modality fusion. One straightforward fusion approach is to directly concatenate visual and audio features without any processing, as used in earlier multimodal VD methods \cite{xd_violence,xd_violence_TMM}. 
However, this approach has clear limitations due to the stark differences between these modalities, which manifest themselves in two key aspects: \textbf{modality information discrepancy} and \textbf{modality asynchrony}.
These two phenomena are particularly prevalent in VD tasks. In many violent events, the audio modality typically captures transient sounds like hits, gunshots, explosions, or screams, while the visual modality conveys richer, more detailed information, such as color variations, facial expressions, and physical interactions. Moreover, the timing of the modality features may differ; for instance, an attack action might precede the victim's scream, or in a shooting incident, audio features might occur earlier than the corresponding visual features, even though both convey the same semantic meaning. Ignoring these modality-specific characteristics and treating both modalities equally can lead to redundant information inclusion and ultimately reduce the effectiveness of the audio modality, even misrepresenting the causal relationships across modalities.

\par
Recently, some methods have attempted to address these two modality discrepancies to achieve more effective modality fusion. To address the issue of modality information discrepancy, a typical approach is to design specialized inter-modal interaction modules for the weighted fusion of modality features, as seen in works like \cite{ICASSP_21_1,MSBT_2024,Pang_TMM}. For the problem of modality asynchrony, Jiashuo Yu et al. \cite{Modality_asynchrony} were the first to tackle this issue, employing audio-to-audio-visual self-distillation to eliminate this discrepancy.
\par
Unlike prior approaches, we argue that the discrepancies between audio and visual modalities fundamentally stem from the distinct roles that vision and hearing play in event perception. Visual modalities typically provide rich spatial and dynamic information, offering a high-dimensional representation of events, whereas audio captures sound characteristics that are often instantaneous and temporally localized. Therefore, in the absence of detailed label information, addressing modality differences at the low-level content is of limited value. Instead, at a higher semantic level, we propose leveraging the inherent properties of these modalities, using their consistency in event representation to align the semantic features of each modality. Multimodal fusion can then be performed based on the aligned semantic features, enabling the full utilization of modality-specific information at the semantic level. Building on this analysis, we introduce a novel multimodal VD framework, which consists of three stages:
\par
\begin{figure*}[!htbp]
	\centering
	\includegraphics[scale=0.23]{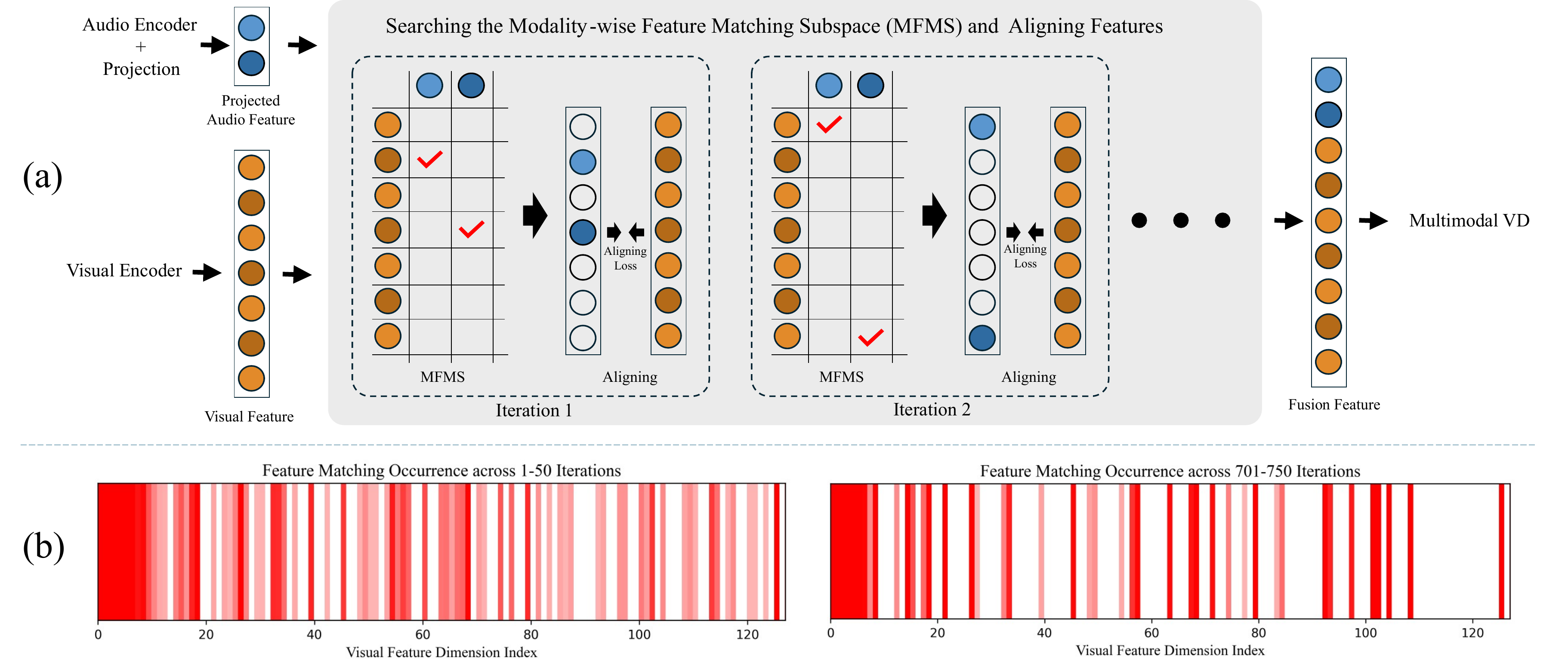}
	\caption{(a) An illustration of Searching for the Modality-wise Feature Matching Subspace(MFMS) and aligning features. In each iteration, we compute the pairwise similarity between audio and visual feature dimensions, select the most matching visual feature dimensions as MFMS. The audio features are mapped into the MFMS, forming sparse features, which are then aligned with the visual features. (b) Visualization of the MFMS convergence process. The red bars represent whether a particular visual feature dimension is identified as part of the MFMS over 50 iterations, with darker colors indicating more frequent identification. It can be observed that at the beginning of the iterations, most dimensions are considered part of the MFMS, and after some iterations, the MFMS converges to a small set of dimensions.}
	\label{fig:sfs}
\end{figure*}

1. \textbf{Unimodal MIL}: This stage focuses on training the encoders for each modality using MIL loss, with the aim of extracting the most relevant semantic features for the VD task.
\par
2. \textbf{Multimodal Alignment}: This stage involves searching for the Modality-wise Feature Matching Subspace (MFMS) and aligning the semantic features based on the identified MFMS. Using the audio-visual search and alignment process as an example (illustrated in Fig. \ref{fig:sfs} (a)). The MFMS is the subspace within the visual semantic feature space that best matches the audio semantic feature space. The existence of the MFMS is based on the assumption that, due to the differing amounts of information provided by audio and visual modalities in event perception, the projection of audio semantic features into the visual semantic feature space should be sparse. Specifically, certain dimensions of the projection should have zero components, indicating that these feature dimensions represent semantic information unique to the visual modality. Non-zero projection components, on the other hand, represent the shared feature space for event expression, which constitutes the MFMS. The search for the MFMS (detailed in Algorithm \ref{alg:1}) is dynamic during training: in the first iteration, we compute pairwise similarity between audio and visual feature dimensions, select the visual feature dimensions that best match as a MFMS. The audio features are then projected into the MFMS, resulting in new sparse audio features, which are then aligned with the visual features. In the next iteration, the MFMS is re-searched, and the audio features are realigned with the visual features. This iterative process continues until it converges to a stable feature subspace, as shown in Fig. \ref{fig:sfs} (b). 

The alignment mentioned above refers to enhancing the consistency of event expression across the two modalities. This involves two aspects: 1) reducing the distance between the semantic features of the two modalities after matching, and 2) ensuring temporal consistency between the modalities. This approach strengthens the modality features that are strongly related to event expression, helping to reduce the impact of redundant information in each modality and improving the complementarity of the modality-specific information.
Our complete model further refines the visual modality into RGB video and optical flow video (referred to as RGB and flow). In the context mentioned earlier, "visual" refers to RGB. Similar to the relationship between audio and RGB, flow focuses on dynamic details, whereas RGB contains more comprehensive information. Therefore, its semantic features also exhibit sparse projection in the RGB feature space. Consequently, the complete model searches for two MFMSs: RGB-Audio and RGB-Flow. The alignment process involves pairwise alignment among the three modalities—audio, RGB, and flow—which will be explained in detail in Section \ref{Method}.
As can be seen, our method essentially establishes a primary modality—RGB. The primary modality constructs the main information framework, with other modalities serving as supplementary within this framework, which aligns with the inherent characteristics of each modality.
\par
3. \textbf{Multimodal Fusion and VD}: Since the previous components have significantly enhanced the "usability" of each modality's features, this part has a simple structure, consisting of an encoder for fusion of modality features and a regressor layer for calculating the violence score. The loss function used for training includes both MIL loss and a specially designed Triplet Loss tailored for this task.
\par
The experimental results on the XD-Violence multimodal VD dataset demonstrate the effectiveness of our method, which improves the average precision (AP) to 86.07\char`\%, significantly surpassing existing related works. The key innovations of our work can be summarized as follows:
\begin{itemize}
	\item We propose a novel method for aligning features at the semantic level by leveraging the inherent properties of each modality. Unlike previous methods, our approach not only overcomes modality differences but also effectively utilizes these differences.
	\item Based on this alignment approach, we introduce a new, simple, and effective framework for multimodal VD.
	\item Experimental results on the XD-Violence dataset demonstrate that the proposed method achieves state-of-the-art performance.
\end{itemize}
\par

\section{Related Work}
\label{}
\subsection{Weakly supervised Violence Detection}
\label{}
Weakly supervised violence detection (VD) aims to identify violent segments using video-level labels, and it is closely related to weakly supervised video anomaly detection. Most existing research treats violent scenes as a subset of anomalous scenes in the context of weakly supervised video anomaly detection. As a result, a significant portion of the methods we refer to as weakly supervised VD are essentially video anomaly detection approaches. Many of these methods treat VD as a purely visual task \cite{MIL_VVD_1, MIL_VVD_2, MIL_VVD_3, MIL_VVD_4, KBS_1, RTFM, MIL_VVD_5, MSL, CVPR_Lv_2023}. While these methods explore various techniques for extracting and processing visual features, they often overlook the potential contributions of other modalities.
\par
Recently, with the release of large-scale audiovisual datasets like XD-Violence \cite{xd_violence}, multimodal weakly supervised VD has gained significant attention. A critical challenge in this domain is how to effectively fuse information from modalities with imbalanced data, especially in the absence of detailed labels. Existing methods \cite{xd_violence}, \cite{xd_violence_TMM}, \cite{Dual_MEM} often merge modality features directly without any preprocessing, which makes them vulnerable to the imbalance between modalities. Some approaches \cite{ICASSP_21_1,MSBT_2024,Pang_TMM} have designed specialized inter-modal interaction modules for weighted fusion of modality features to mitigate these issues. Jiashuo Yu et al. \cite{Modality_asynchrony} were the first to address the problem of modality asynchrony between audio and visual data, employing audio-to-audio-visual self-distillation to eliminate this discrepancy.
In contrast to previous approaches, we leverage the inherent imbalances across modalities and design a novel multimodal semantic feature alignment method. This method aims to improve the utilization of each modality's information by aligning their semantic features, thus enhancing the effectiveness of multimodal fusion in VD.
\subsection{Multimodal Alignment and Fusion}
\label{}
Alignment and fusion are core concepts in multimodal learning \cite{ali_fus_1,Multimodal_deep_learning}. Although they are distinct, they are complementary and interdependent \cite{ali_fus, ali_fus_1}. The goal of multimodal alignment is to address the heterogeneity between different modalities (such as text, images, audio, video, etc.) by establishing semantic consistency, so that these modalities can express similar or related semantic information within a common representation space. Multimodal fusion, on the other hand, is the process of effectively integrating information from multiple modalities to generate a unified representation \cite{ali_fus, ali_fus_1, ali_fus_54}. Recent studies suggest that performing alignment before fusion enhances the fusion process. This is because alignment ensures that data from different sources are synchronized in terms of time, space, or context, making their combination meaningful \cite{ali_fus_49}. Alignment also allows less informative modalities to be effectively utilized \cite{ali_fus_107}, and ensures that the relationships between different modalities are well understood and accurately modeled \cite{ali_fus_1, ali_fus_53}. These factors contribute to the ability of the multimodal fusion process, based on aligned data, to capture more comprehensive and useful information \cite{ali_fus}. In our approach, multimodal alignment is primarily designed to facilitate multimodal fusion. Experimental results demonstrate that once the modalities are aligned, a straightforward fusion strategy can achieve excellent performance.
\par
The literature \cite{ali_fus_1} classifies multimodal alignment into two types: Explicit Alignment and Implicit Alignment. A key characteristic of Explicit Alignment is the direct measurement of similarity, while Implicit Alignment generally does not explicitly align the data; instead, alignment is a part of the model's latent structure in tasks involving modality interactions, such as Image Captioning \cite{Image_Captioning}, Visual Question Answering \cite{Visual_Question_Answering}, etc.
From this perspective, our alignment method is closer to Explicit Alignment. The foundational work in Explicit Alignment includes Canonical Correlation Analysis (CCA) \cite{CCA} and its deep learning extension, Deep CCA \cite{Deep_CCA}. CCA finds a linear transformation that projects data from two modalities into a shared space and maximizes their correlation, while Deep CCA uses deep neural networks for nonlinear mapping, allowing for better alignment of features across modalities. In recent multimodal learning works, CCA is often used as a loss function to align modalities \cite{CCA_1, CCA_2, CCA_3}.
Our approach differs from previous methods. The similarity measure aims to identify subspaces in the feature space of the primary modality (the main source of information) that correspond to the secondary modality (which provides complementary information) in the event representation. These subspaces are a part of the primary modality's feature space and are used to sparsely map the features of the secondary modality into the primary modality's feature space.The alignment loss used for training (see Section \ref{3-2}) primarily considers the features of the primary modality and sparsified the secondary modality, their pairwise distances, and the consistency of the mapping of modality features in the temporal decision sequence space.
\par

\section{Method}
\label{Method}
\begin{figure*}[!htbp]
	\centering
	\includegraphics[scale=0.18]{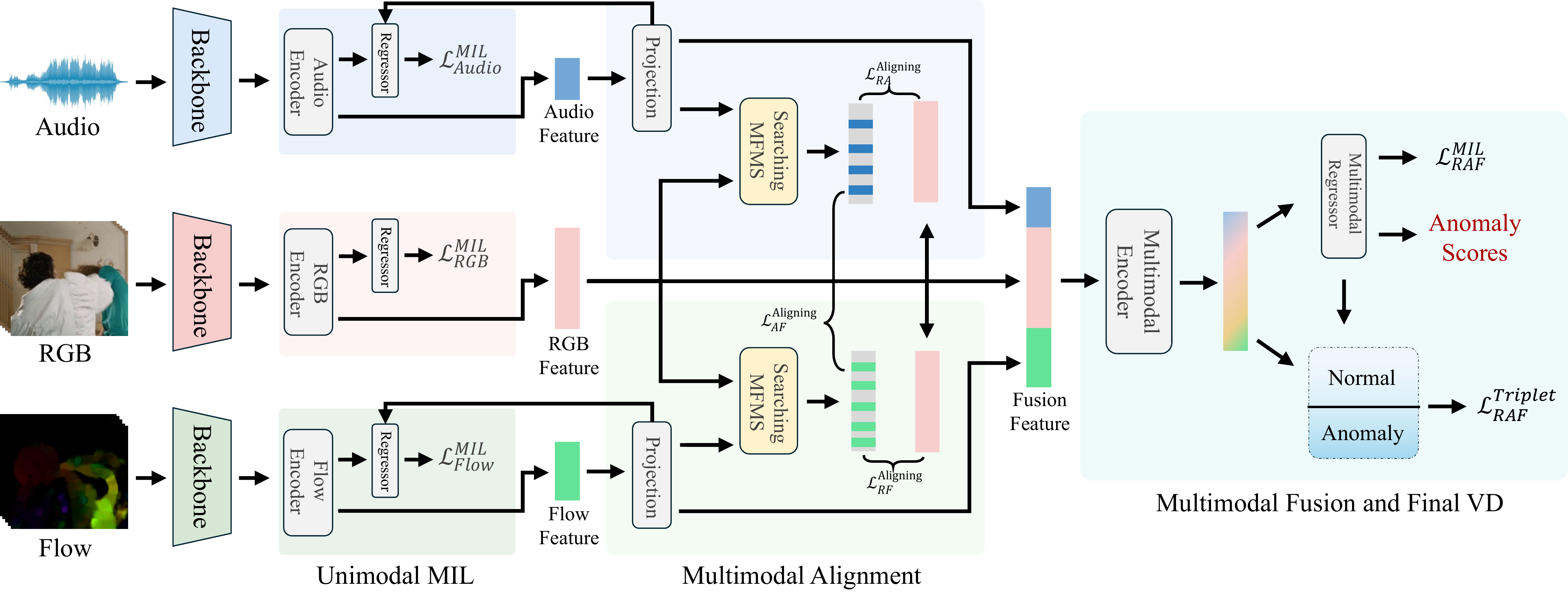}
	\caption{An overview of the proposed framework. It includes three stages: 1. Unimodal MIL, this stage focuses on training the encoders for each modality using MIL loss, with the aim of extracting the most relevant semantic features for the VD task. 2. Multimodal alignement, in this stage, our proposed method searches for the MFMSs and aligns the semantic features of different modalities based on the identified MFMSs. 3. Multimodal Fusion and final VD, this stage utilizes a multimodal encoder to fuse the aligned modality features and trains the model using both MIL loss and a specially designed Triplet Loss tailored for the VD task. }
	\label{fig:overall}
\end{figure*}

In this paper, we propose a novel weakly supervised multimodal VD framework that efficiently leverages information from three distinct modalities: RGB, audio, and flow. This framework includes unimodal semantic feature extraction, multimodal alignment, and fusion processes. The overall architecture is illustrated in Fig. \ref{fig:overall}. Below, we provide a detailed description of our methodology.

\subsection{Unimodal Semantic Feature Extraction}
\label{}
Our modality alignment is performed at the semantic feature level. Prior to alignment, it is essential to extract the semantic features relevant to the VD task from each modality. Therefore, this stage is designed to train the encoders to extract the most pertinent semantic features for each modality in the context of the VD task.

Each modality's encoder takes as input the visual or auditory feature extracted from pre-trained networks (e.g., I3D \cite{I3D} for RGB and flow, and VGGish \cite{vgg_1,vgg_2} for audio). \( f_R \), \( f_F \), and \( f_A \) represent the features extracted by pre-trained networks for RGB, flow, and audio, respectively. The structure of the encoder is the same for each modality, with the only difference being the input and output feature dimensions. The encoder consists of a single 1D convolutional layer followed by a transformer module.

The 1D convolutional layers are used to extract local temporal features, and the features from different modalities are reduced to varying dimensions based on the amount of information each modality contains:

\begin{align}
	f_R^c &\in \mathbb{R}^{T \times D_R} = \text{Conv1d}_R(f_R), \\
	f_F^c &\in \mathbb{R}^{T \times D_F} = \text{Conv1d}_F(f_F), \\
	f_A^c &\in \mathbb{R}^{T \times D_A} = \text{Conv1d}_A(f_A).
\end{align}

Where \( D_R \), \( D_F \), and \( D_A \) are the feature dimensions of the different modalities. Based on the amount of information each modality contains, we assume \( D_R > D_F > D_A \).

Each layer of the Transformer module consists of Global and Local Multi-Head Self-Attention (GL-MHSA) and a Feed Forward Network (FFN), designed to extract semantic features that incorporate both local and global information. The GL-MHSA, proposed by \cite{Dual_MEM}, introduces a local temporal mask in addition to global attention, enabling the model to capture both long-range dependencies and local structures. This enhances the model’s expressive power when handling complex sequential data. The FFN further processes these features, providing nonlinear transformations to improve the model's fitting capacity, while Layer Normalization (LN) ensures the stability of the training process. The Transformer module’s \( l \)-th layer can be formulated as follows:

\begin{align}
	&\hat{ z }^l = \text{LN}(\text{GL-MHSA}(z^l)) + z^l, \\
	&z^{l+1} = \text{LN}(\text{FFN}(\hat{ z }^l)) + \hat{ z }^l.
\end{align}

Where, \( z^l \) represents the input features of the \( l \)-th layer. After passing through all the Transformer layers, we obtain the high-level features for each modality: \( z_R \), \( z_F \), and \( z_A \). 

To ensure that the high-level features are strongly correlated with the VD task, we treat the training of each modality's encoder as a weakly supervised VD task. Specifically, we first apply a regression layer at the end of each modality's encoder to obtain frame-level anomaly scores. This regression layer consists of a three-layer MLP, which computes the anomaly score for each time step:

\begin{align}
	s_R = \text{Regressor}_R(z_R),\\
	s_F = \text{Regressor}_F(z_F),\\
	s_A = \text{Regressor}_A(z_A).
\end{align}

Due to the lack of frame-level annotations, we adopt a MIL loss, which employs the widely-used top-K strategy: it averages the top-K anomaly scores, i.e., \( \bar{s} = \frac{1}{K} \sum_{i \in T_K(s)} s_i \), where \( T_K(s) \) represents the set of top-K scores in \( s \). Thus, the MIL loss for each modality can be described as follows:
\begin{align}
	\mathcal{L}_R^{MIL} = -y \log(\bar{s}_R) - (1 - y) \log(1 - \bar{s}_R),\\
	\mathcal{L}_F^{MIL} = -y \log(\bar{s}_F) - (1 - y) \log(1 - \bar{s}_F),\\
	\mathcal{L}_A^{MIL} = -y \log(\bar{s}_A) - (1 - y) \log(1 - \bar{s}_A).
\end{align}
Here, \(y\) is the video-level label.
\par
\par
\subsection{Multimodal Alignment}
\label{3-2}
The goal of this stage is to enhance modality-specific features that are closely related to event representation through modality alignment, ensuring semantic consistency across modalities and improving the complementarity of cross-modal features. 

As described in Section \ref{s1}, our alignment is achieved based on the search for MFMSs, which are the subspaces in the primary modality's semantic feature space most relevant to the secondary modality in event representation. The choice of primary modality is based on the following analysis: Compared to the RGB modality, audio captures sound features related to violent events that are often instantaneous and temporally localized, meaning the projection of audio semantic features into the visual semantic feature space should be sparse. Similarly, the flow modality, compared to RGB, focuses on dynamic details, while RGB contains more comprehensive information; thus, its projection in the RGB feature space should also be sparse. Consequently, we treat RGB as the primary modality and identify two MFMSs within the RGB modality's feature space—one for audio and one for flow.

Before identifying the MFMSs, we fix the RGB semantic features and process the audio and flow semantic features through separate projection layers, each consisting of three MLP layers. The rationale behind this approach is that RGB, as the primary modality, provides the main structural information, while audio and flow serve as supplementary modalities, embedded within this primary structure to offer local details. Therefore, these two modalities need to actively align with the RGB modality. The projection layers offer an additional structure, built on the unimodal semantic feature extraction models, to facilitate this active alignment. The process of projecting audio and flow features can be described as follows:

\begin{align}
	\hat{ z_A } = \text{Projection}_A(z_A),\\
	\hat{ z_F } = \text{Projection}_F(z_F).
\end{align}

The process of searching for the MFMS is shown in Algorithm \ref{alg:1}. First, a similarity matrix between the feature dimensions of the primary and secondary modalities is computed based on the information from the current batch size. Then, based on this similarity matrix, we identify which dimensions in the high-dimensional space of the primary modality correspond to feature dimensions in the low-dimensional space of the secondary modality. After identifying the corresponding modality dimensions, the features of the secondary modality are embedded into the feature space of the primary modality, thus forming a new sparse secondary modality feature. This process can be expressed concisely as:
\begin{align}
	\tilde{z_A} = \text{Sparse}_A(\hat{ z_A } \,|\, \text{MFMS of } z_R \text{ for } \hat{ z_A }),\\
	\tilde{z_F} = \text{Sparse}_F(\hat{ z_F }\,|\, \text{MFMS of } z_R \text{ for } \hat{ z_F }).
\end{align}

After obtaining the sparse audio and flow features, the first alignment can be performed. This alignment process is achieved by minimizing three losses, which are designed to increase the pairwise similarity between the three modalities:
\begin{align}
	\mathcal{L}_\text{RA}^{\text{Cos}} = 1 - \text{CosineSimilarity}(z_R, \tilde{z_A}),\\
	\mathcal{L}_\text{RF}^{\text{Cos}} = 1 - \text{CosineSimilarity}(z_R, \tilde{z_F}),\\
	\mathcal{L}_\text{AF}^{\text{Cos}} = 1 - \text{CosineSimilarity}(\tilde{z_A}, \tilde{z_F}).
\end{align}
Where the CosineSimilarity between \( x \) and \( y \) is defined as:
\begin{align}
	\text{CosineSimilarity}(x, y) = \frac{x \cdot y}{\|x\|_2 \|y\|_2}.
\end{align}

\begin{figure}[!htbp]
	\centering
	\includegraphics[scale=0.3]{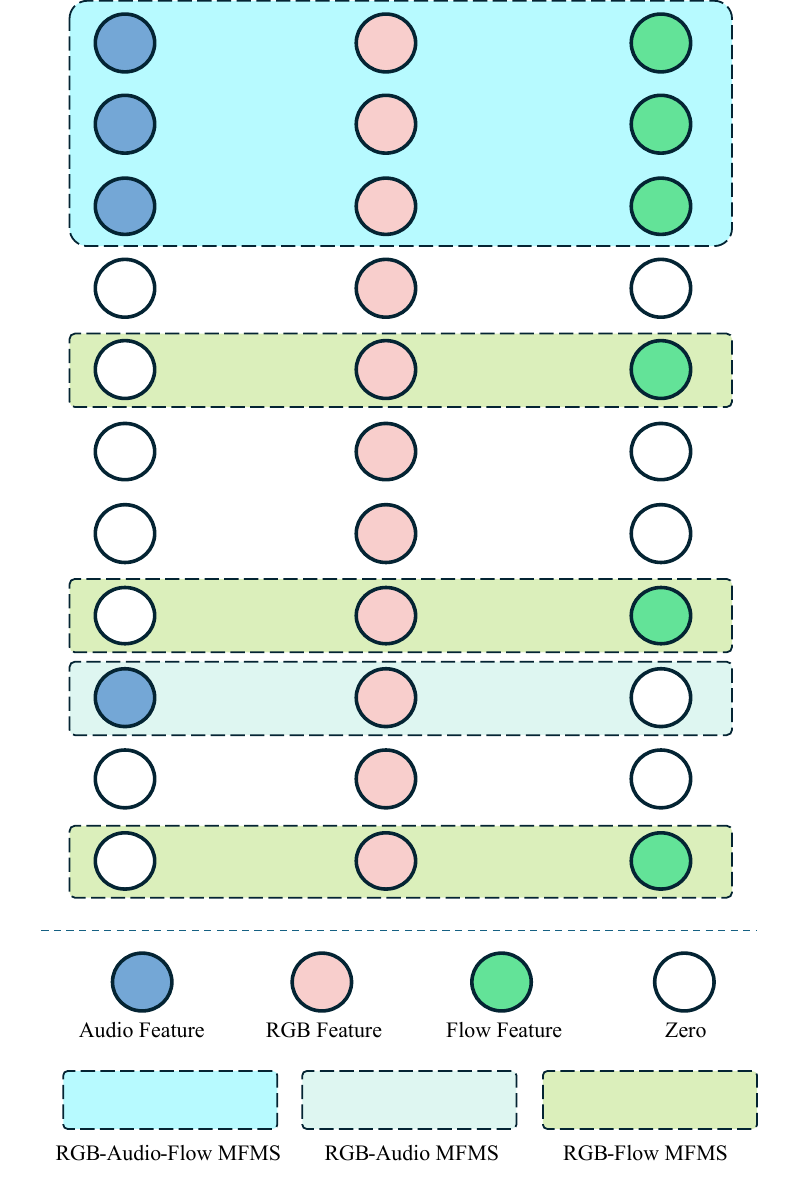}
	\caption{By searching for MFMSs, the entire RGB feature space is divided into four distinct parts: RGB-Audio-Flow MFMS, RGB-Audio MFMS, RGB-Flow MFMS, and pure RGB.}
	\label{fig:3-MFMSs}
\end{figure}

Following the search for the MFMS and embedding of the features, as shown in Fig. \ref{fig:3-MFMSs}, the entire RGB feature space is divided into four parts: RGB-Audio-Flow MFMS, RGB-Audio MFMS, RGB-Flow MFMS, and pure RGB. This suggests that certain semantic information in the event is jointly expressed by all three modalities, some is expressed by both RGB and audio or flow, and some is only expressed by the RGB modality. The three losses mentioned above essentially bring closer the features that are jointly expressed across multiple modalities, enhancing the relationships between different modalities when representing the same event. This maximizes shared information and helps the fusion model better understand and correlate complementary information from different modalities, thus reducing the impact of redundant information. From another perspective, only the most strongly correlated dimensions of the modalities are brought closer by these losses, which helps the model find the optimal MFMSs during training iterations.

  \begin{algorithm}
	\caption{Searching for MFMS and sparsifying secondary modality features based on MFMS.}
	\label{alg:1}
	\begin{algorithmic}[1]
		\Statex \textbf{Input:} Secondary modality feature $z_s \in \mathbb{R}^{b \times t \times d_s}$
		\Statex \hspace{1.3cm} Primary modality feature $z_p \in \mathbb{R}^{b \times t \times d_p}$
		\Statex $\texttt{\#}$  Where $b$ and $t$ represent the batch size and temporal dimension, $d_s$ and $d_p$ represent feature dimension, $ d_s < d_p $.		 
		\Statex \textbf{Output:}  
		Sparse secondary modality feature $\tilde{z_s} \in \mathbb{R}^{b \times t \times d_p}$ based on MFMS.
		
		\State 
		Define \(
		\hat{z_s} \in \mathbb{R}^{(b \times t) \times d_s}, \quad \hat{z_p} \in \mathbb{R}^{(b \times t) \times d_p}.
		\) \hspace{0.5cm} $\texttt{\#}$ Flatten the inputs into 2D.
		
		\State Normalize each column of \( \hat{z_s} \) and \( \hat{z_p} \): 
		\(
		\hat{z_s}[i, j] = \frac{\hat{z_s}[i, j]}{\| \hat{z_s}[:, j] \|_2}, 
		\hat{z_p}[i, j] = \frac{\hat{z_p}[i, j]}{\| \hat{z_p}[:, j] \|_2}.
		\)
		
		\State Compute the normalized similarity matrix $S$ between $\hat{z_s}$ and $\hat{z_p}$:
		\[
		S = \hat{z_s}^\top \hat{z_p} \in \mathbb{R}^{d_s \times d_p}.
		\]
		
		\State Initialize \( S_{\text{top-k}} \in \mathbb{N}^{d_s \times d_p} \) as an empty matrix, and \( \theta \in \mathbb{N}^{d_s} \) as an empty vector. 
		\hspace{1.1cm} $\texttt{\#}$ The $\theta$ is used for storing the dimension indices of MFMS.
		\State $used\_values \gets \emptyset$  \hspace{3cm} $\texttt{\#}$ Used to track used feature dimension indexs.
		
	    \For{$i = 1 \text{ to } d_s$}
   			\State $ S_{\text{top-k}}[i, :] = \text{argsort}(S[i, :])[:k]$  \hspace{0.5cm} $\texttt{\#}$ Where, $\text{argsort}(S[i, :])$ returns the indices of the sorted elements of $S[i, :]$, with $ d_s \le k\le d_p$.
			\For{$j = 1 \text{ to } k$}
				\State $candidate \gets S_{\text{top-k}}[i, j]$
				\If{$candidate \notin \text{used\_values}$}
					\State $\theta[i] \gets candidate$
					\State $\text{used\_values} \gets \text{used\_values} \cup \{candidate\}$
					\State \textbf{break}
				\EndIf
			\EndFor
		\EndFor
		
\State Define 
\(
\hat{\theta} = \{ i \in \{1, 2, \dots, d_p\} \mid i \notin \theta \}
\). 
\hspace{0.2cm} $\texttt{\#}$ \( \hat{\theta} \) as the set of feature dimension indices from \( 1 \) to \( d_p \) that are not in \( \theta \).

\State Define 
\(
\theta^{\text{pad}} = [\theta, \hat{\theta}]
\).
\hspace{3cm} $\texttt{\#}$ Concatenating \( \theta \) and \( \hat{\theta} \).

\State Create a zero tensor 
\(
O \in \mathbb{R}^{b \times t \times (d_p - d_s)} 
\).

\State Expand the tensor: 
\(
z_s^{\text{pad}} = [z_s, O]
\).

\State Rearrange the elements of the expanded tensor according to the expanded index set \( \theta^{\text{pad}} \) to obtain the sparse secondary modality feature:
\[
\tilde{z_s} = z_s^{\text{pad}}[:, :, \theta^{\text{pad}}]
\]
	\end{algorithmic}
\end{algorithm}

Additionally, we apply constraints to the audio features \( \hat{z_A} \) and flow features \( \hat{z_F} \) after the projection layer to ensure the alignment of modal features in the temporal sequence. This can be regarded as the second stage of alignment.

Firstly, to ensure that the features after the projection layer do not lose event-related semantic information, a regression operation is performed. The regression layers are reused from the previous stage, and the results are constrained by the MIL loss, as described below:
\begin{align}
	\hat{s_A} = \text{Regressor}_A(\hat{z_A}),
	\quad \hat{\mathcal{L}}_A^{\text{MIL}} = \text{MIL}(\hat{s_A}, y).\\
	\hat{s_F} = \text{Regressor}_F(\hat{z_F}),
	\quad \hat{\mathcal{L}}_F^{\text{MIL}} = \text{MIL}(\hat{s_F}, y).
\end{align}

Secondly, we align the anomaly score sequences of each modality, namely \( \hat{s_A} \), \( \hat{s_F} \), and \( s_R \) (the anomaly score sequence of the RGB modality, which remains the same as in the previous stage), by minimizing the pairwise differences in the temporal sequence using three loss functions:

\begin{align}
	\mathcal{L}_{RA}^{S-\text{CE}} &= \text{Score-CrossEntropy}(s_R, \hat{s_A}), \label{eq:loss_ra} \\
	\mathcal{L}_{RF}^{S-\text{CE}} &= \text{Score-CrossEntropy}(s_R, \hat{s_F}), \label{eq:loss_rf} \\
	\mathcal{L}_{AF}^{S-\text{CE}} &= \text{Score-CrossEntropy}(\hat{s_A}, \hat{s_F}). \label{eq:loss_af}
\end{align}
Where the Score-CrossEntropy  between \( \mathbf{p} \) and \( \mathbf{q} \) is defined as:
\begin{align}
	\text{Score-CrossEntropy}(\mathbf{p}, \mathbf{q}) &= 
	- \frac{1}{N} \sum_{i=1}^{N} \Big[ 
	\text{clamp}(p_i, \epsilon, 1 - \epsilon) 
	\log \left( \text{clamp}(q_i, \epsilon, 1 - \epsilon) \right) \nonumber \\
	&\quad + \left( 1 - \text{clamp}(p_i, \epsilon, 1 - \epsilon) \right) 
	\log \left( 1 - \text{clamp}(q_i, \epsilon, 1 - \epsilon) \right) \Big]. 
\end{align}
Where \( p_i \) and \( q_i \) are the \( i \)-th elements of vectors \( \mathbf{p} \) and \( \mathbf{q} \), respectively, and \( N \) is the number of elements in the vectors. The clamp operation is defined as:
\begin{align}
\text{clamp}(x, \epsilon, 1 - \epsilon) = \max(\epsilon, \min(x, 1 - \epsilon)),
\end{align}
where \( \epsilon \) is a small constant to prevent issues with taking the logarithm of zero.

Eq. \eqref{eq:loss_ra}, \eqref{eq:loss_rf}, and \eqref{eq:loss_af} define the loss function that minimizes discrepancies between anomaly score sequences from different modalities, enabling the projection model to capture cross-modal causal relationships. In simpler terms, it identifies features susceptible to modality asynchrony and applies adaptive "compensation" to mitigate inconsistencies.

Based on the above discussion, the modality alignment loss can be defined as pairwise alignment among the three modalities:
\begin{align}
	\mathcal{L}_{RAF}^{MA} &= \mathcal{L}_{RA}^{Aligning} + \mathcal{L}_{RF}^{Aligning} + \mathcal{L}_{AF}^{Aligning}.
	\label{eq:modality_alignment_loss} 
\end{align}
Where,
\begin{align}
	\mathcal{L}_{RA}^{Aligning} &= \mathcal{L}_{RA}^{Cos} + \mathcal{L}_{RA}^{S-CE}, \\
	\mathcal{L}_{RF}^{Aligning} &= \mathcal{L}_{RF}^{Cos} + \mathcal{L}_{RF}^{S-CE}, \\
	\mathcal{L}_{AF}^{Aligning} &= \mathcal{L}_{AF}^{Cos} + \mathcal{L}_{AF}^{S-CE} + \lambda \cdot (\hat{\mathcal{L}}_A^{\text{MIL}} + \hat{\mathcal{L}}_F^{\text{MIL}}).
\end{align}
Since $\hat{\mathcal{L}}_A^{\text{MIL}}$ and $\hat{\mathcal{L}}_F^{\text{MIL}}$ serve only as auxiliary terms, we set $\lambda$ to 0.01.

\par
\subsection{Multimodal Fusion and Violence Detection }
\label{}
\par
Due to the significant enhancement in the usability of each modality's features after alignment, this part of the model remains relatively simple. It consists of a multimodal encoder for fusing the modality features and a regression layer for calculating final anomaly scores.

The input to the multimodal encoder is the newly fused feature, which is formed by concatenating the aligned features of each modality:
\begin{align}
	z_{RAF} = [\hat{z_A} \| z_R \| \hat{z_F}].
\end{align}

The multimodal encoder consists of a Linear layer and a Temporal Convolutional Network (TCN) block. The Linear layer is composed of two MLP layers and is responsible for weighted fusion of the modality features into a unified representation. The TCN block is employed to further merge the features at a higher level, capturing temporal dependencies and more abstract patterns. This process can be described as:
\begin{align}
	\hat{z_{RAF}} = \text{TCN}(\text{Linear}(z_{RAF})).
\end{align}
\par
There are two important points that require further clarification:
1) \textbf{The non-use of sparse features \( \tilde{z_A} \) and \( \tilde{z_F} \):} These features are not sufficiently "compact" and may interfere with the fusion model. Moreover, during the inference stage, using them would increase the model's complexity. Their role is completed during the modality alignment process, where they have already served their purpose.
2) \textbf{The use of TCN blocks instead of Transformers:} While Transformers are better suited for handling long-term dependencies, in our framework, the single-modality part already addresses long-term dependencies and global feature extraction. The alignment stage also considers the full duration of the modalities for alignment. However, in the final fusion stage, we do not need to focus heavily on global information. Instead, the task is to output frame-level predictions, which are more closely related to local features. Furthermore, TCNs are more efficient in this context, as they reduce the number of parameters and improve computational efficiency.
\par
The regression layer of this stage is also composed of three MLP layers. The process of calculating the anomaly score sequence by the regression layer can be expressed as:
\begin{align}
	s_{RAF} = \text{Regressor}_{RAF}(\hat{z_{RAF}}).
\end{align}
The MIL Loss in this case also adopts the top-K mode, which can be simply expressed as:
\begin{align}
	\mathcal{L}_{RAF}^{MIL} = \text{MIL}_{\text{top-K}}(s_{RAF}, y).
\end{align}
\par
Additionally, in order to enhance the discrimination of fused features in abnormal samples, we designed a triplet loss \cite{Triplet_loss} for this task, as described in Algorithm \ref{alg:2}. Based on the video-level labels, all anomaly score sequences and multimodal fused features within a batch are divided into two classes: normal and anomaly.

In the triplet loss, the anchor is defined as the mean of the features corresponding to the top-k anomaly scores in the normal class. The positive is defined as the mean of the features corresponding to the smallest k anomaly scores in the anomaly class, while the negative is defined as the mean of the features corresponding to the largest k anomaly scores in the anomaly class. This formulation ensures that the anchor (normal class) is pulled closer to the positive (smallest anomaly scores in the anomaly class), while maintaining a large distance from the negative (largest anomaly scores in the anomaly class).

The process of computing the triplet loss can be simply expressed as:

\begin{align}
	\mathcal{L}_{RAF}^{Triplet} = \Psi(\hat{z_{RAF}}, s_{RAF}, y),
\end{align}
\noindent where $\Psi$ represents the process of calculating the triplet loss.

\begin{algorithm}
	\caption{Triplet Loss Calculation}
	\label{alg:2}
	\begin{algorithmic}[1]
		\Statex \textbf{Input:} Fusion features: $\hat{z_{RAF}} \in \mathbb{R}^{b \times t \times d}$
		\Statex \hspace{1.3cm} Score sequence: $s_{RAF} \in \mathbb{R}^{b \times t}$
		\Statex \hspace{1.3cm} Image-level label: $y \in \{0, 1\}^b$		 
		\Statex \textbf{Output:} $\mathcal{L}_{RAF}^{Triplet}$
		
		\Statex $\texttt{\#}$ Get normal sets and anomaly sets
		\State $N_{\text{label}} \gets \{ i \mid y[i] = 0, \, i = 1, 2, \dots, b \}$
		\State $A_{\text{label}} \gets \{ i \mid y[i] = 1, \, i = 1, 2, \dots, b \}$
		\State Define $b_N$  and $b_A$ is the number of elements in $N_{\text{label}}$ and $A_{\text{label}}$.  
		\If{$b_N = 0$ \textbf{or} $b_A = 0$ }
			\State \Return 0
		\EndIf
		\State \( N_{\text{feature}} \gets  \hat{z_{RAF}} [N_{\text{label}}, :, :] \), \( N_{\text{score}} \gets s_{RAF}[N_{\text{label}}, :]  \).	
		\State \( A_{\text{feature}} \gets \hat{z_{RAF}} [A_{\text{label}} , :, :]  \), \( A_{\text{score}} \gets  s_{RAF}[A_{\text{label}}, :] \).
		
		\Statex $\texttt{\#}$ Get anchor feature
		\State Initialize \( anchor \in \mathbb{R}^{b_N \times d} \) as an zero matrix.
		\For{$i = 1$ to $b_N$}
		\State $\text{index} \gets $ the top-k largest indices from $N_{\text{score}}[i, :]$
		\State $tmp \gets N_{\text{feature}}[i, \text{index}, :]$
		\State $anchor[i, :] \gets \frac{1}{k} \sum_{j=1}^{k} tmp[j, :]$
		\EndFor
		\State $f_\text{anchor} \gets \frac{1}{b_N} \sum_{i=1}^{b_N} anchor[i,:]$
		
		\Statex $\texttt{\#}$ Get positive feature and negative feature
		\State Initialize \( positive \in \mathbb{R}^{b_A \times d} \)  and \( negative \in \mathbb{R}^{b_A \times d} \) as zero matrixs.
		\For{$i = 1$ to $b_A$}
		\State $\text{index} \gets \text{the top-k smallest indices from}$ $A_{\text{score}}[i,:]$
		\State $tmp \gets A_{\text{feature}}[i, \text{index}, :]$
		\State $positive[i, :] \gets \frac{1}{k} \sum_{j=1}^{k}tmp[j, :]$
		\State $\text{index} \gets \text{the top-k largest indices from}$ $A_{\text{score}}[i,:]$
		\State $tmp \gets A_{\text{feature}}[i, \text{index}, :]$
		\State $negative[i, :] \gets \frac{1}{k} \sum_{j=1}^{k}tmp[j, :]$
		\EndFor
		\State $f_\text{positive} \gets \frac{1}{b_A} \sum_{i=1}^{b_A} positive[i, :]$, 
		$f_\text{negative} \gets \frac{1}{b_A} \sum_{i=1}^{b_A} negative[i, :]$.
		
		\Statex $\texttt{\#}$ Compute Triplet Margin Loss
		\State $\mathcal{L}_{RAF}^{Triplet} \gets \text{triplet}(\text{norm}(f_\text{anchor}), \text{norm}(f_\text{positive}), \text{norm}(f_\text{negative}))$
		\State \Return  $\mathcal{L}_{RAF}^{Triplet}$
	\end{algorithmic}
\end{algorithm}

\subsection{Loss Functions}
\label{}
Based on the previous content, our loss function consists of four components:

\begin{itemize}
	\item The unimodal MIL loss for extracting semantic features from each modality:
	\(
	\mathcal{L}_{RAF}^{U-MIL} = \mathcal{L}_R^{MIL} + \mathcal{L}_A^{MIL} + \mathcal{L}_F^{MIL}.
	\)
	\item The modality alignment loss:
	\(
	\mathcal{L}_{RAF}^{MA}.
	\)
	\item The multimodal MIL loss:
	\(
	\mathcal{L}_{RAF}^{M-MIL}.
	\)
	\item The triplet loss for enhancing the discriminative power of fused features for anomaly samples:
	\(
	\mathcal{L}_{RAF}^{Triplet}.
	\)
\end{itemize}
Thus, the total loss function during training can be expressed as:
\begin{align}
\mathcal{L} = \mathcal{L}_{RAF}^{U-MIL} 
+ \lambda_1 \cdot \mathcal{L}_{RAF}^{MA}
+ \lambda_2 \cdot \mathcal{L}_{RAF}^{M-MIL} 
+ \lambda_3 \cdot \mathcal{L}_{RAF}^{Triplet}.
\end{align}
Where \(\lambda_1\), \(\lambda_2\) and \(\lambda_3\) are the hyperparameters that control the relative importance of each loss component.
\subsection{Inference}
\label{}

\begin{figure*}[!htbp]
	\centering
	\includegraphics[scale=0.15]{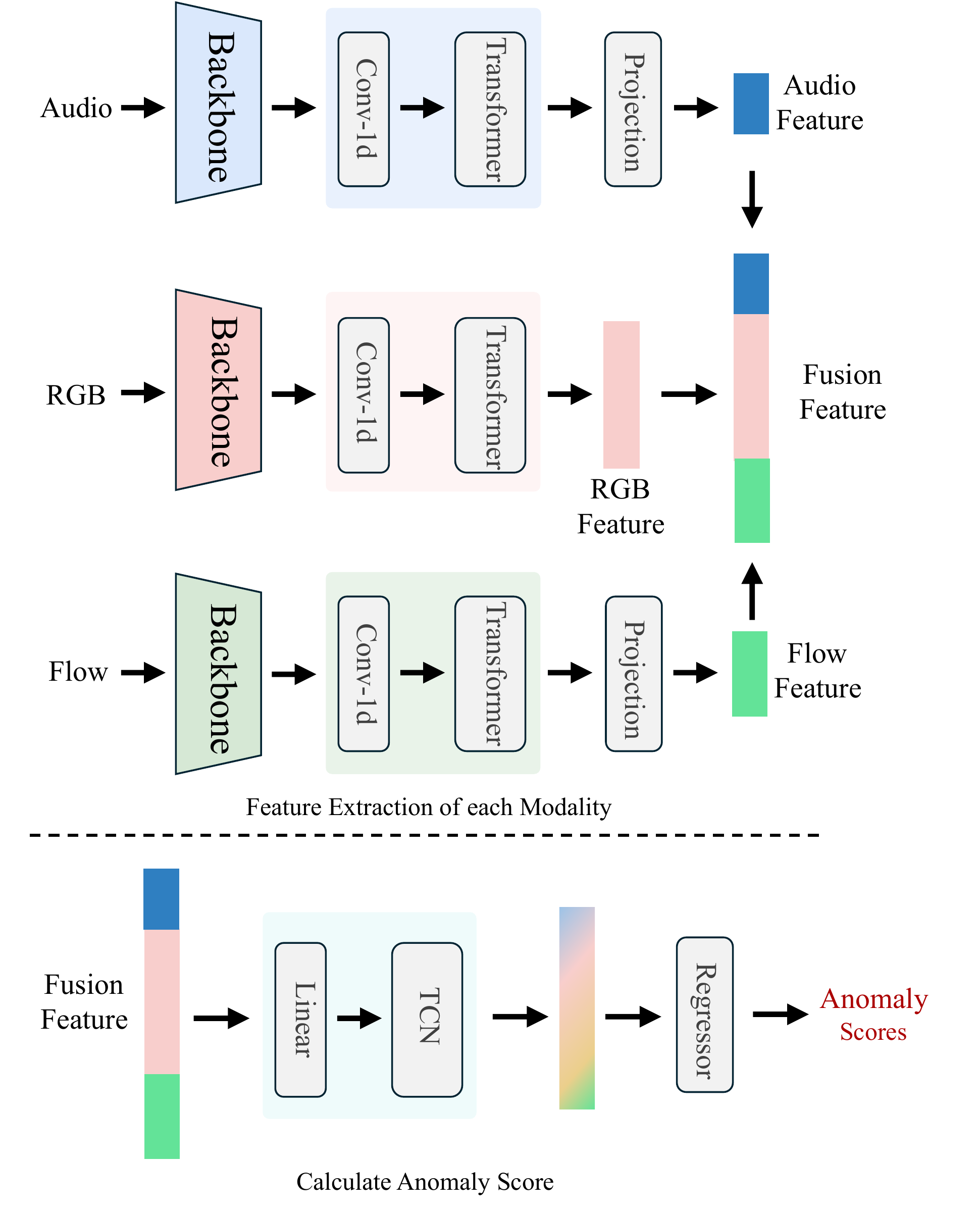}
	\caption{Overview of our inference process. The process comprises two stages: Stage 1 Feature Extraction of each Modality. Stage 2, Fusion multimodal featrues and calculate violent score.}
	\label{fig:infer}
\end{figure*}

The inference phase is depicted in Fig. \ref{fig:infer}. Its structure is quite simple and comprises two stages: Stage 1: Feature Extraction for Each Modality, and Stage 2: Fusion of Multimodal Features and Calculation of Anomaly Score for VD.

In Stage 1, the output is the aligned features from each modality. We have made the model parameters for this part publicly available, so future research can easily build upon our model and explore other multimodal fusion methods based on the aligned features.

\par

\section{Experiment}
\label{}
In this section, we comprehensively evaluate the performance of our method through comparisons with state-of-the-art approaches and ablation experiments targeting different modules. The experimental details and analysis are presented below.
\subsection{Dataset and Evaluation Metric}
\label{}
The XD-Violence \cite{xd_violence} dataset is the only large-scale publicly available multimodal dataset for VD to date, containing 4,754 untrimmed videos with a total duration of 217 hours. The training set consists of 3,954 videos, including 1,905 violent and 2,049 non-violent videos, while the test set includes 800 videos, with 500 violent and 300 non-violent videos. The videos in XD-Violence come from various scenes, including movies and YouTube, and cover a range of common violence types, giving the dataset significant advantages in terms of diversity and realism. Each video provides three modalities: RGB video, audio, and flow, offering rich resources for multimodal VD. 

Since the XD-Violence test set provides frame-wise annotation labels for the videos, we follow prior works \cite{xd_violence_TMM, Modality_asynchrony, Xiao2022, MSBT_2024} and use frame-level average precision (AP) as the evaluation metric during inference.
\par
\subsection{Implementation Details}
\label{trainsets}
Consistent with existing methods \cite{xd_violence_TMM, Modality_asynchrony,Xiao2022,MSBT_2024}, the backbone for extracting RGB and flow features is the I3D \cite{I3D} network pre-trained on the Kinetics-400 dataset, while the backbone for audio feature extraction is the VGGish \cite{vgg_1,vgg_2} network pre-trained on the large YouTube dataset. In the visual feature extraction process, the frame rate for all videos is fixed at 24 FPS, and the sliding window length is set to 16 frames. The audio is divided into overlapping 960-millisecond segments, with each audio segment uniquely corresponding to a visual segment. The log-mel spectrogram blocks of size 96 × 64 bins computed from each segment serve as the input to the VGGish network.

Our method is implemented based on PyTorch. In the process of extracting unimodal semantic features, the encoders for RGB, audio, and flow modalities use 1D convolutional layers to reduce the feature dimensions extracted by the backbone to 128, 32, and 64, respectively, with each modality's Transformer block containing 4 attention heads and 2 layers. In the multimodal fusion stage, the TCN in the multimodal encoder has a 3-layer structure, with an output feature dimension of 64. During training, the Top-K value for all modalities is computed as \( L/16 + 1 \), where \( L \) is the sequence length in the sample. We use a fixed learning rate of 0.0001, $\text{betas} = (0.9, 0.999), \text{weight decay} = 0.0005$, and a batch size of 128 with the Adam optimizer for 1000 training iterations. The hyperparameters \( (\lambda_1, \lambda_2, \lambda_3 \) are set to \( (10.0, 10, 0.001) \) to balance the total loss. Our experiments were conducted on a computer equipped with an Intel i7 11700T CPU (16M Cache, up to 4.60 GHz), 64GB of RAM, and an Nvidia RTX A4000 GPU.

\par
\subsection{Comparisons with State-of-the-Arts on XD-Violence Dateset}
\label{}

\begin{table}[htbp]
	\centering
	\caption{Performance of Different Methods on XD-Violence Dataset}
	\renewcommand{\arraystretch}{1.2} 
	\begin{tabular}{@{} l c c @{}} 
		\toprule 
		\textbf{Method} & \textbf{Modality} & \textbf{AP (\%)} \\ 
		\toprule
		\multicolumn{3}{l}{\textit{\textbf{Unsupervised Learning-Based Methods}}} \\
		\cmidrule(lr){1-3} 
		SVM baseline & RGB & 50.78 \\
		OCSVM \cite{unsup_1} & RGB & 27.25 \\
		Hasan et al. \cite{unsup_2} & RGB & 30.77 \\
		\midrule
		\multicolumn{3}{l}{\textit{\textbf{Weakly Supervised Learning-Based Methods}}} \\
		\cmidrule(lr){1-3} 
		Sultani et al. \cite{Sultan} & RGB & 75.68 \\
		Wu et al. \cite{MIL_VVD_3} & RGB & 75.9 \\
		RTFM \cite{RTFM} & RGB & 77.81 \\
		MSL \cite{MSL} & RGB & 78.28 \\
		HL-Net \cite{xd_violence} & RGB + Audio & 78.64 \\
		ACF \cite{ACF} & RGB + Audio & 80.13 \\
		Pang et al. \cite{Pang_TMM} & RGB + Audio & 79.37 \\
		Zhang et al. \cite{Zhang_2023_CVPR} & RGB + Audio & 81.43 \\
		MACIL-SD (Light) \cite{Modality_asynchrony} & RGB + Audio & 82.17 \\
		MACIL-SD (Full) \cite{Modality_asynchrony} & RGB + Audio & 83.4 \\
		\textbf{HyperVD} \cite{HyperVD} & \textbf{RGB + Audio} & \textbf{85.67} \\
		MSBT \cite{MSBT_2024} & RGB + Audio & 82.54 \\
		MSBT \cite{MSBT_2024} & RGB + Flow & 80.68 \\
		Wu et al. \cite{xd_violence_TMM} & RGB + Audio + Flow & 79.53 \\
		Xiao et al. \cite{Xiao2022} & RGB + Audio + Flow & 83.09 \\
		MSBT \cite{MSBT_2024} & RGB + Audio + Flow & 84.32 \\
		\midrule
		\rowcolor{gray!10} 
		Ours & RGB + Audio & 85.15 \\
		\rowcolor{gray!10} 
		\textbf{Ours} & \textbf{RGB + Flow} & \textbf{84.59} \\
		\rowcolor{gray!10} 
		\textbf{Ours} & \textbf{RGB + Audio + Flow} & \textbf{86.07} \\
		\bottomrule 
	\end{tabular}
	\label{tab:cp}
\end{table}

Table \ref{tab:cp} presents the performance of various methods on the XD-Violence dataset. The table categorizes the methods into two main groups: unsupervised learning-based methods and weakly supervised learning-based methods.

In the category of unsupervised learning-based methods, the SVM baseline achieves the highest performance with an AP of 50.78$\%$, while other methods, such as OCSVM \cite{unsup_1} and Hasan et al. \cite{unsup_2}, perform relatively lower, with APs of 27.25$\%$ and 30.77$\%$, respectively. These results highlight the challenge of relying solely on unsupervised learning for VD in videos. In the absence of labels, models struggle to distinguish between violent and non-violent content based solely on learning data structures.

The weakly supervised learning-based methods can be further classified based on the number of modalities utilized: unimodal weakly supervised methods, methods using two modalities, and methods that can simultaneously utilize three modalities.

Unimodal weakly supervised methods include works by Sultani et al. \cite{Sultan}, Wu et al. \cite{MIL_VVD_3}, RTFM \cite{RTFM}, and MSL \cite{MSL}. Methods that use two modalities are typically those combining RGB and audio features, such as HL-Net \cite{xd_violence}, ACF \cite{ACF}, Zhang et al. \cite{Zhang_2023_CVPR}, MACIL-SD \cite{Modality_asynchrony}, and HyperVD\cite{HyperVD} which achieve a maximum AP of 85.69$\%$. This demonstrates the significant role audio features play in enhancing VD performance. Additionally, multimodal methods combining RGB and flow, such as MSBT \cite{MSBT_2024}, show an AP of 80.68$\%$, which, while slightly lower than methods using audio, still represents a significant improvement over unimodal methods.

There are relatively few methods that combine three modalities. Some methods, such as MACIL-SD \cite{Modality_asynchrony}, which perform exceptionally well with two modalities, are difficult to extend to three-modal applications due to their unique audio-visual self-distillation structure. Among existing three-modality methods, Wu et al. \cite{xd_violence_TMM} employ an early fusion structure, while Xiao et al. \cite{Xiao2022} and MSBT \cite{MSBT_2024} adopt a weighted fusion of modality features. Notably, MSBT achieves an AP of 84.32$\%$, significantly outperforming other methods.

Our model can operate with both two-modality and three-modality inputs. For example, with two-modality inputs—RGB and audio or RGB and flow—the main modality is RGB, with audio and flow serving as supplementary modalities. In these cases, the detection results are 85.19$\%$ and 84.59$\%$, respectively. When utilizing all three modalities, the performance improves further, with an AP of 86.07$\%$, surpassing all the methods in the table.

\par
\subsection{Training Generalization, Robustness, and Inference Efficiency}
\label{}

\begin{table}[htbp]
	\centering
	\caption{Performance Comparison on Different Training Set Proportions and Test Sets}
	\label{tab:3trainset}
	\renewcommand{\arraystretch}{1.5} 
	\begin{tabular}{>{\centering\arraybackslash}p{3cm}|cc|cc|cc}
		\toprule
		\multirow{2}{*}{ \textbf{Method}} & \multicolumn{2}{c|}{30\% Training Set} & \multicolumn{2}{c|}{70\% Training Set} & \multicolumn{2}{c}{100\% Training Set} \\
		\cline{2-7}
		& Test A & Test B & Test A & Test B & Test A & Test B \\
		\midrule
		MACIL-SD \cite{Modality_asynchrony} & 80.54 & 77.32 & 82.88 & 79.95 & 84.73 & 81.87 \\
		HyperVD \cite{HyperVD} & 80.39 & 76.42 & 82.17 & 77.94 & 86.93 & 84.07 \\
		Wu et al. \cite{xd_violence_TMM} & 79.76 & 74.07 & 80.15 & 75.95 & 82.21 & 76.69 \\
		MSBT \cite{MSBT_2024} & 82.52 & 77.63 & 81.07 & 78.79 & 85.48 & 82.97 \\
		\rowcolor{gray!5} 
		Ours & \textbf{83.56} & \textbf{82.80} & \textbf{85.09} & \textbf{83.03} & \textbf{87.28} & \textbf{84.52} \\
		\bottomrule
	\end{tabular}
	
	\vspace{0.2cm}
	\raggedright
	\scriptsize\emph{Note}: All values are AP (\%). Test A/B denote different test subsets.
\end{table}

To evaluate the generalization ability and robustness of our approach, we designed a series of comparative experiments to investigate the accuracy performance of different methods under various training set ratios and testing subsets. Additionally, we analyzed the performance variations of each method when applying different frame drop rates during the training phase. The comparative methods include two typical tri-modal fusion strategies: the direct concatenation fusion method Wu et al. \cite{xd_violence_TMM} and the weighted fusion method MSBT \cite{MSBT_2024}, as well as a unique dual-modal fusion method based on self-distillation, namely MACIL-SD \cite{Modality_asynchrony}.

As shown in Table \ref{tab:3trainset}, we randomly selected different proportions (30\% and 70\%) of samples from the XD-Violence training set to form new training sets, and equally divided the testing set into two subsets, Test A and Test B. Based on this configuration, we compared the accuracy performance of various methods under different training and testing conditions to validate their generalization ability across different data distributions. Experimental results demonstrate that the performance of weakly supervised multimodal violence detection methods is positively correlated with the scale of the training dataset; as the amount of training data increases, the performance of all methods improves. Under the 30\% training set condition, our method achieves APs of 83.56\% and 82.80\% on Test A and Test B, respectively, significantly outperforming the competing methods. When the training set is increased to 70\%, our method further raises the AP to 85.09\% (Test A) and 83.03\% (Test B), which remains markedly higher than that of other methods. This fully demonstrates the superior generalization ability of our approach across different data distributions.

\begin{table}[htbp]
	\centering
	\caption{Performance Comparison under Varying Training-phase Frame Drop Rates}
	\label{tab:drop}
	\renewcommand{\arraystretch}{1.3}
	\setlength{\tabcolsep}{12pt} 
	\begin{tabular}{c|cccc}
		\toprule
		\textbf{Method} & No Drop & 10\% Drop & 30\% Drop & 50\% Drop \\ 
		\midrule
		MACIL-SD \cite{Modality_asynchrony} & 83.40 & 80.47 & 78.46 & 77.57 \\
		HyperVD \cite{HyperVD} & 85.67 & 78.67 & 78.42 & 76.99 \\
		Wu et al. \cite{xd_violence_TMM} & 79.53 & 78.41 & 78.57 & 77.09 \\ 
		MSBT \cite{MSBT_2024} & 84.32 & 79.42 & 80.06 & 74.96 \\
		\rowcolor{gray!5} 
		Ours & \textbf{86.07} & \textbf{84.30} & \textbf{81.56} & \textbf{78.22} \\ 
		\bottomrule
	\end{tabular}
	
	\vspace{0.2cm}
	\raggedright
	\scriptsize\emph{Note}: All values are AP in percentage. Frame drop rates are applied only during training phase, while evaluation uses complete test samples.
\end{table}

We applied different frame drop rates (10\%, 30\%, and 50\%) to the modal data of each sample in the XD-Violence training set to construct new datasets that simulate modality dropout in real-world scenarios, and then observed the performance variations of different methods. As shown in Table \ref{tab:drop}, although the AP of all methods decreases with increasing frame drop rates, our method consistently outperforms the alternatives. Notably, under 10\% and 30\% frame drop conditions, our method achieves APs of 84.30\% and 81.56\%, respectively (while other methods yield APs ranging from 78.41\% to 80.47\%). Even under the extreme condition of a 50\% frame drop rate, our method maintains an AP of 78.22\%, surpassing the performance of the other methods. This fully verifies the robustness of our approach in handling incomplete inputs.

\begin{table}[htbp]
	\centering
	\caption{Comparative Analysis of Model Complexity and Inference Efficiency Across Three Typical Three-Modal Fusion Methods on XD-Violence Benchmark}
	\label{tab:param-infer-compare}
	\begin{tabular}{c|ccc}
		\toprule
		\textbf{Method} & Params (M) & Inf. Time (s/sample) & AP (\%) \\
		\midrule
		Wu et al.~\cite{xd_violence_TMM} & 1.43 & \textbf{0.05} & 79.53 \\
		MSBT~\cite{MSBT_2024} & 15.7 & 0.19 & 84.32 \\
		Ours & \textbf{1.36} & 0.07 & \textbf{86.07} \\
		\bottomrule
	\end{tabular}
\end{table}

Table~\ref{tab:param-infer-compare} presents a comparative analysis of model complexity and inference efficiency for three typical three-modal fusion methods on the XD-Violence benchmark. The table reports the number of parameters (in millions), the average inference time per sample (in seconds), and the detection accuracy (AP\%).
It can be seen that the direct concatenation fusion method by Wu et al. benefits from its simple structure, achieving the fastest inference speed and a relatively low parameter count, though its detection performance is limited. In contrast, the MSBT method, with its complex modal weighted fusion mechanism, effectively enhances detection accuracy—albeit at the cost of increased model complexity and slower inference. By comparison, our method achieves the highest detection accuracy while maintaining a low overall parameter count and fast inference speed.

\par


\par
\subsection{Ablation Study}
\label{Ablation experiments}

Table \ref{tab:loss} shows the impact of each component in the loss function on the results during training. The presence or absence of each loss term also indicates whether the corresponding module is fully utilized. The presence of $\mathcal{L}_{RAF}^{U-MIL}$ determines whether the features extracted by each modality's encoder are strongly correlated semantic features for the VD task. The presence of $\mathcal{L}_{RAF}^{MA}$ indicates whether modality alignment is performed. The term $\mathcal{L}_{RAF}^{MIL}$ directly affects the final regression layer, determining whether the model is adapted to the VD task. Finally, $\mathcal{L}_{RAF}^{Triplet}$ aims to enhance the discrimination of fused features for abnormal samples.

\begin{table}[htbp]
	\centering
	\caption{Performance of Different Loss Functions in the Training}
	\renewcommand{\arraystretch}{1.2} 
	\begin{tabular}{@{} c | c c c c c | c @{}} 
		\toprule 
		\textbf{Index} & \textbf{$\mathcal{L}_{RAF}^{U-MIL}$} & \textbf{$\mathcal{L}_{RAF}^{MA}$} & \textbf{$\mathcal{L}_{RAF}^{MIL}$ } & \textbf{$\mathcal{L}_{RAF}^{Triplet}$} & \textbf{AP (\%)} \\
		\midrule 
		1 & & \checkmark  & \checkmark & \checkmark & 82.67 \\
		2 & \checkmark &  & \checkmark & \checkmark & 84.03 \\
		3 & \checkmark & \checkmark  & & \checkmark & 83.80 \\
		4 & \checkmark & \checkmark & \checkmark & & 85.91 \\
		5 & & & \checkmark & \checkmark & 81.82 \\
		6 & \checkmark & \checkmark & \checkmark & \checkmark & \textbf{86.07} \\
		\bottomrule 
	\end{tabular}%
	\label{tab:loss}
\end{table}

\begin{figure}[!htbp]
	\centering
	\includegraphics[scale=0.6]{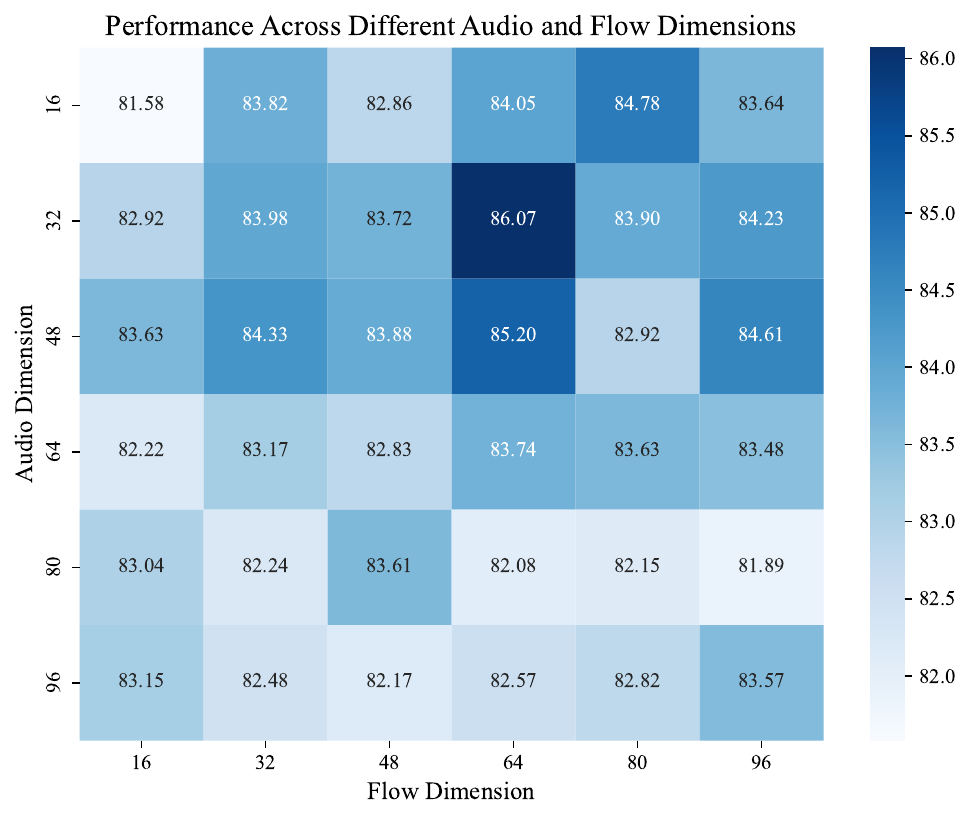} 
	\caption{Performance across different audio and flow dimensions. The AP ($\%$) values are visualized as a heatmap, where deeper blue represents higher AP values.}
	\label{fig:vfdim}
\end{figure}

In Table \ref{tab:loss}, the result in the row with Index 1 reflects the case when $\mathcal{L}_{RAF}^{U-MIL}$ is missing, showing a significant decrease in the AP (\%) value compared to when it is included. In this case, the features extracted by the unimodal encoders are not strongly correlated semantic features for the VD task. As previously described, our modality alignment module is designed based on the consistency of semantic features across modalities in event representation. Without this, the alignment module's role becomes unclear, which in turn negatively impacts the overall detection performance.
The result in the row with Index 2 reflects the detection performance when the modality alignment loss $\mathcal{L}_{RAF}^{MA}$ is absent. This also leads to a noticeable drop in detection performance, highlighting that modality alignment improves modality fusion effectiveness.
The result in the row with Index 3 underscores the importance of $\mathcal{L}_{RAF}^{MIL}$, while also indicating that $\mathcal{L}_{RAF}^{Triplet}$ can serve as a constraint for the VD task's regression layer. The result in Index 4 shows a slight decrease in detection performance when $\mathcal{L}_{RAF}^{Triplet}$ is absent. This can be attributed to the fact that the XD-Violence dataset contains a considerable number of abnormal video samples where the entire sequence is violent, diminishing the impact of $\mathcal{L}_{RAF}^{Triplet}$.
The result in Index 5 represents the case where only the final detection loss is present, making the model a simple late fusion multi-modal VD model. However, this approach performs significantly better compared to the early fusion method proposed by Wu et al. \cite{xd_violence_TMM}.

Fig. \ref{fig:vfdim} presents the performance of our method on the XD-Violence dataset when audio and flow features are reduced to different dimensionalities. The heatmap in the figure shows the AP ($\%$) values for each combination of audio dimensions (16, 32, 48, 64, 80, 96) and flow dimensions (16, 32, 48, 64, 80, 96). The color intensity represents the corresponding AP value, with darker colors indicating higher AP values, which suggests better model performance for those dimension combinations.
From the figure, it can be observed that as the audio and flow dimensions change, the model’s performance varies across different combinations. Notably, in certain combinations of audio and flow dimensions, the AP value shows a significant increase. For example, when the audio dimension is 32 and the flow dimension is 64, the model achieves the highest AP value of 86.07$\%$. This result indicates that certain combinations of audio and flow features in specific dimensions can significantly improve the model's performance, and these dimensions reflect the combinations that minimize redundancy in audio and flow semantic features in event representation. It is worth noting that in our method, these dimensionalities are manually set, and future research could explore how to automatically identify the optimal dimensionalities.

\setlength{\tabcolsep}{12pt}  
\begin{table}[htbp]
	\centering
	\caption{Ablation Study of Hyperparameters}
	\label{tab:param-compare}
	\begin{tabular}{@{}ccccc@{}}
		\toprule
		\textbf{Index} & \textbf{$\lambda_1$} & \textbf{$\lambda_2$} & \textbf{$\lambda_3$} & \textbf{AP (\%)} \\
		\midrule
		1  & \multirow{2}{*}{1}  & \multirow{2}{*}{1}  & 0.001 & 85.73  \\
		2  &                     &                     & 0.01  & 85.82  \\
		\cmidrule(lr){1-5}
		3  & \multirow{2}{*}{1}  & \multirow{2}{*}{5}  & 0.001 & 85.54  \\
		4  &                     &                     & 0.01  & 85.71   \\
		\cmidrule(lr){1-5}
		5  & \multirow{2}{*}{1}  & \multirow{2}{*}{10} & 0.001 & 85.42  \\
		6  &                     &                     & 0.01  & 85.46  \\
		\cmidrule(lr){1-5}
		7  & \multirow{2}{*}{10} & \multirow{2}{*}{1}  & 0.001 & 86.06  \\
		8  &                     &                     & 0.01  & 86.00 \\
		\cmidrule(lr){1-5}
		9  & \multirow{2}{*}{10} & \multirow{2}{*}{5}  & 0.001 & 85.99 \\
		10 &                     &                     & 0.01  & 85.97 \\
		\cmidrule(lr){1-5}
		11 & \multirow{2}{*}{10} & \multirow{2}{*}{10} & 0.001 & \textbf{86.07}\\
		12 &                     &                     & 0.01  & 85.92  \\
		\bottomrule
	\end{tabular}
\end{table}

Table~\ref{tab:param-compare} presents the results of the ablation study on hyperparameters $\lambda_1$, $\lambda_2$, and $\lambda_3$, where the performance is measured by AP (\%). 
$\lambda_1$ is the coefficient of $\mathcal{L}_{RAF}^{MIL}$, which is strongly correlated with the final output. Since it plays a crucial role in optimizing the primary objective, we examine whether setting it to 1 or 10 affects the results. 
$\lambda_2$ is the coefficient of $\mathcal{L}_{RAF}^{MA}$, which is closely related to the alignment function. To explore the impact of alignment weight on performance, we consider values of 1, 5, and 10. 
$\lambda_3$ is the coefficient of $\mathcal{L}_{RAF}^{Triplet}$, which mainly enhances the model's ability to distinguish violent content from non-violent content within violent samples. As it serves as an auxiliary term, we assign it a relatively small weight of 0.001 or 0.01.
The results indicate that different combinations of $\lambda_1$ and $\lambda_2$ lead to slight variations in AP. The highest AP (86.07\%) is achieved when $\lambda_1 = 10$, $\lambda_2 = 10$, and $\lambda_3 = 0.001$, suggesting that this setting is optimal among the tested configurations.

\subsection{Qualitative Analysis}
\label{}

\begin{figure}[!htbp]
	\centering
	\includegraphics[width=\linewidth]{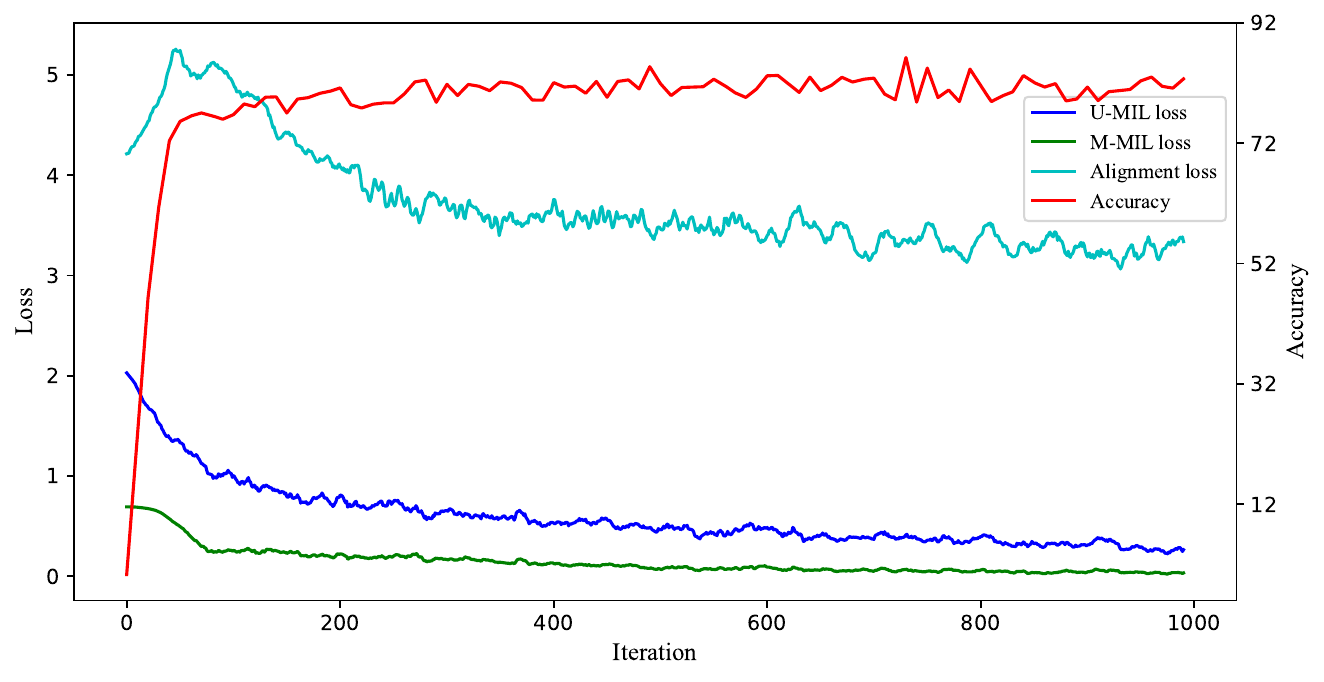} 
	\caption{Loss and Accuracy Curves during Training. The red curve represents the frame-level prediction accuracy. The blue, green, and cyan curves correspond to the Unimodal MIL loss, Multimodal MIL loss, and Modality Alignment loss, respectively, which are most strongly related to the results.}
	\label{fig:loss_ap}
\end{figure}

Fig. \ref{fig:loss_ap} illustrates the evolution of multiple loss functions and accuracy across training iterations. The red curve represents the frame-level prediction accuracy. The blue, green, and cyan curves correspond to the unimodal MIL loss, multimodal MIL loss, and modality alignment loss, respectively, which are most strongly correlated with the results. As observed, in the early stages of training, both the unimodal MIL loss and multimodal MIL loss decrease rapidly. During this phase, the model primarily focuses on extracting the most relevant semantic features for the VD task from each modality, while the alignment module, which is involved in the preparation phase (such as finding the MFMS), shows minimal progress. After approximately 50 iterations, the alignment loss begins to decrease rapidly, and the improvement in accuracy is mainly driven by the alignment process, ultimately achieving the best accuracy.

\begin{figure}[htbp]
	\centering
	\subfloat[MFMS Convergence Metrics over Iterations]{\includegraphics[width=0.95\textwidth]{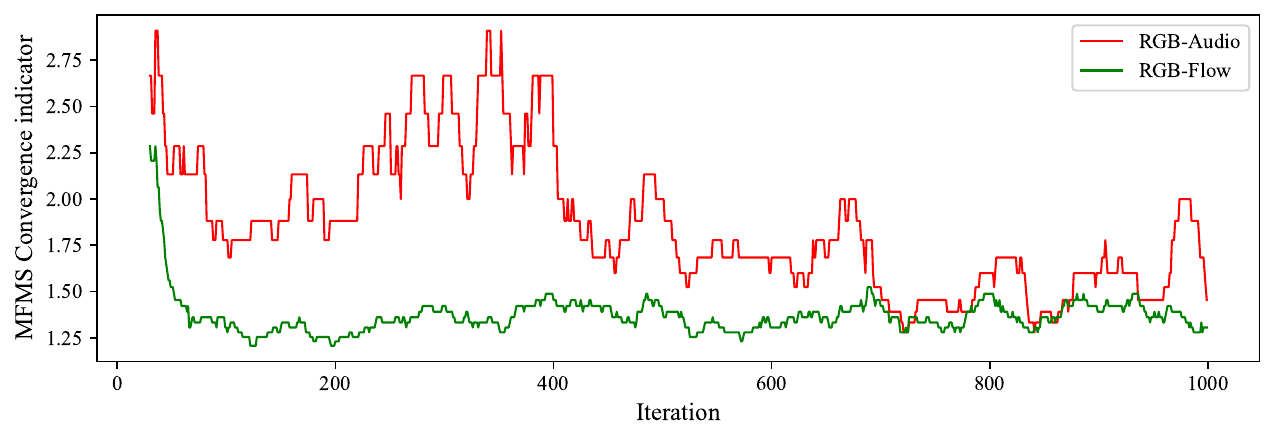}} \\
	\subfloat[RGB-Audio MFMS Distribution]{\includegraphics[width=0.45\textwidth]{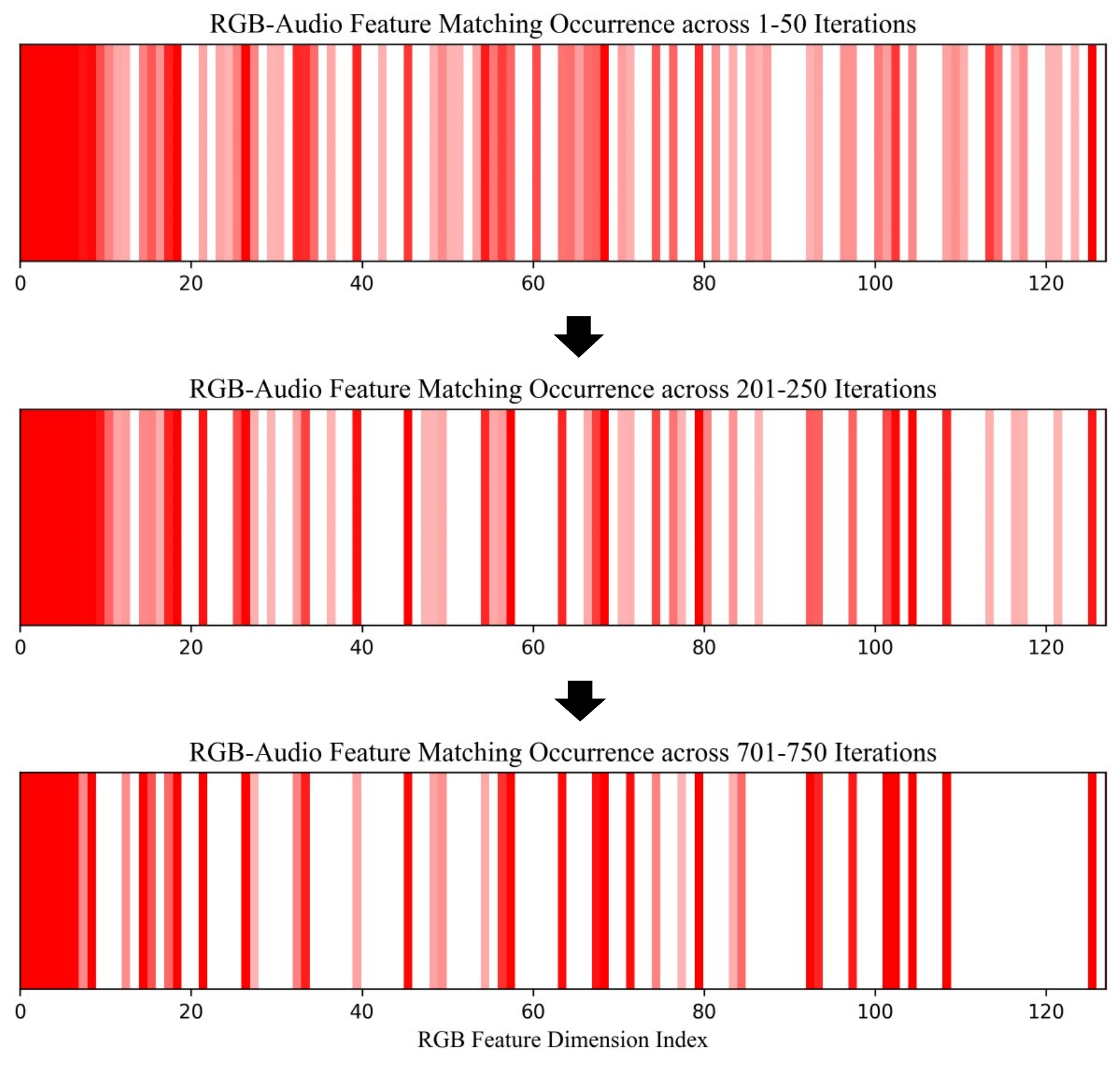}}
	\hspace{0.01\textwidth}
	\subfloat[RGB-Flow MFMS Distribution]{\includegraphics[width=0.45\textwidth]{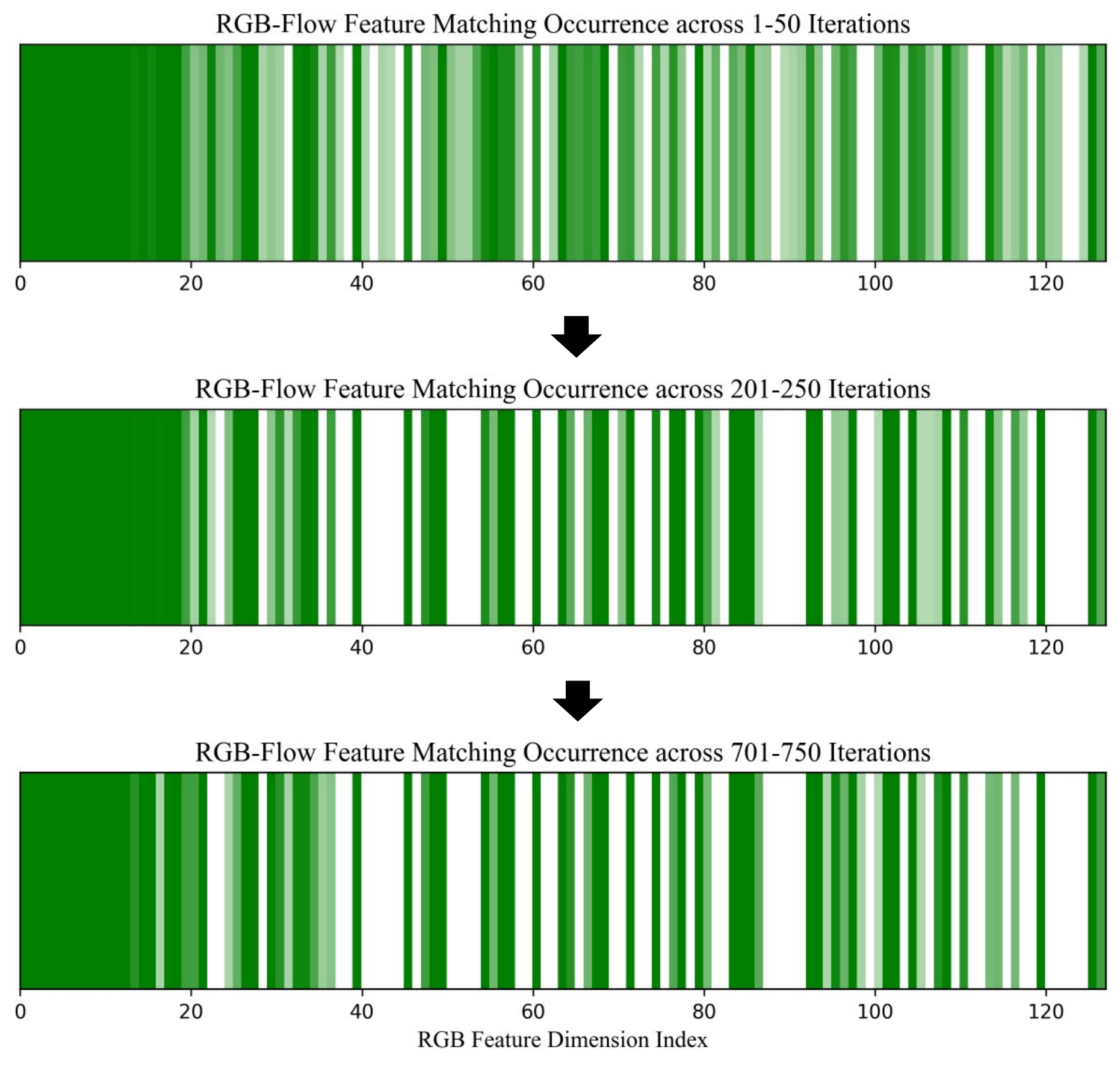}}
	\caption{Visualization of the convergence process for RGB-Audio and RGB-Flow MFMS. In (a), the plot shows the evolution of the MFMS convergence indicator during iterations, with the red curve representing RGB-Audio and the green curve representing RGB-Flow. In (b), red bars indicate the frequency with which each RGB feature dimension is selected as part of the RGB-Audio MFMS over 50 iterations—the darker the red, the more frequently it is chosen. Similarly, in (c), green bars depict the frequency with which each RGB feature dimension is identified as part of the RGB-Flow MFMS over 50 iterations, with darker green corresponding to higher frequency.}
	\label{fig:triangular}
\end{figure}

Fig. \ref{fig:triangular} visualizes the convergence process for the RGB-Audio and RGB-Flow MFMS. In (a), the plot illustrates the evolution of the MFMS convergence indicator over iterations. The MFMS convergence indicator at the \(i\)th iteration are computed as follows:

First, a sliding window of length \(w\) is defined. The set of MFMS results within this window is given by
\(
\{M_{i-w}, M_{i-(w-1)}, \ldots, M_{i}\},
\)
which represents MFMSs identified over the last \(w\) iterations starting from iteration \(i\).

Let \(d_p\) denote the dimensionality of the primary modality’s feature space. Within the sliding window, the number of times the \(k\)th feature (with \(1\le k\le d_p\)) of the primary modality is recognized as part of the MFMS is computed as
\begin{equation}
f^{k}_i = \sum_{j=i-w}^{i} \mathbf{1}_{\{k \in M_j\}},
\end{equation}
where \(\mathbf{1}_{\{k \in M_j\}}\) is the indicator function that equals 1 if \(k \in M_j\) and 0 otherwise.

Using these counts, we determine the maximum count \(f^{\max}_i\) among all features and count the number of features \(n_i\) that reach this maximum:
\begin{align}
	f^{\max}_{i} &= \max_{1\le k \le d_p} f^{k}_i, \label{eq1} \\
	n_i &= \sum_{k=1}^{d_p} \delta(f^{k}_i - f^{\max}_i), \label{eq2}
\end{align}
with \(\delta(x)\) denoting the Kronecker delta function.

The MFMS convergence indicator capture two aspects: if \(f^{\max}_{i}\) is closer to the window length \(w\), an MFMS is more likely to exist; and if \(n_i\) is closer to the secondary modality’s feature dimension \(d_s\), the distribution of the MFMS is more stable. Therefore, the MFMS convergence indicator at the \(i\)th iteration is computed by
\begin{equation}
	\text{m}_i = \frac{w \cdot d_s}{f^{\max}_i \cdot n_i}. \label{eq3}
\end{equation}

In (a), the red and green curves represent the convergence indicator for the RGB-Audio and RGB-Flow MFMS, respectively, clearly indicating that the RGB-Flow MFMS converges faster than the RGB-Audio MFMS.
(b) and (c) show the changes in the distribution of the RGB-Audio and RGB-Flow MFMS in the RGB feature space during training, further demonstrating that the RGB-Flow MFMS converges more rapidly.
This is because the Flow modality can be derived directly from the RGB modality, making it easier to identify the MFMS in the RGB space, whereas the Audio and RGB modalities originate from different sources and have significantly different data structures, resulting in a slower process for identifying the RGB-Audio MFMS in the RGB space.

\begin{figure}[htbp]
	\centering
	\begin{minipage}{0.31\textwidth}
		\centering
		\scriptsize \textbf{(i)} 
		\includegraphics[width=0.95\linewidth]{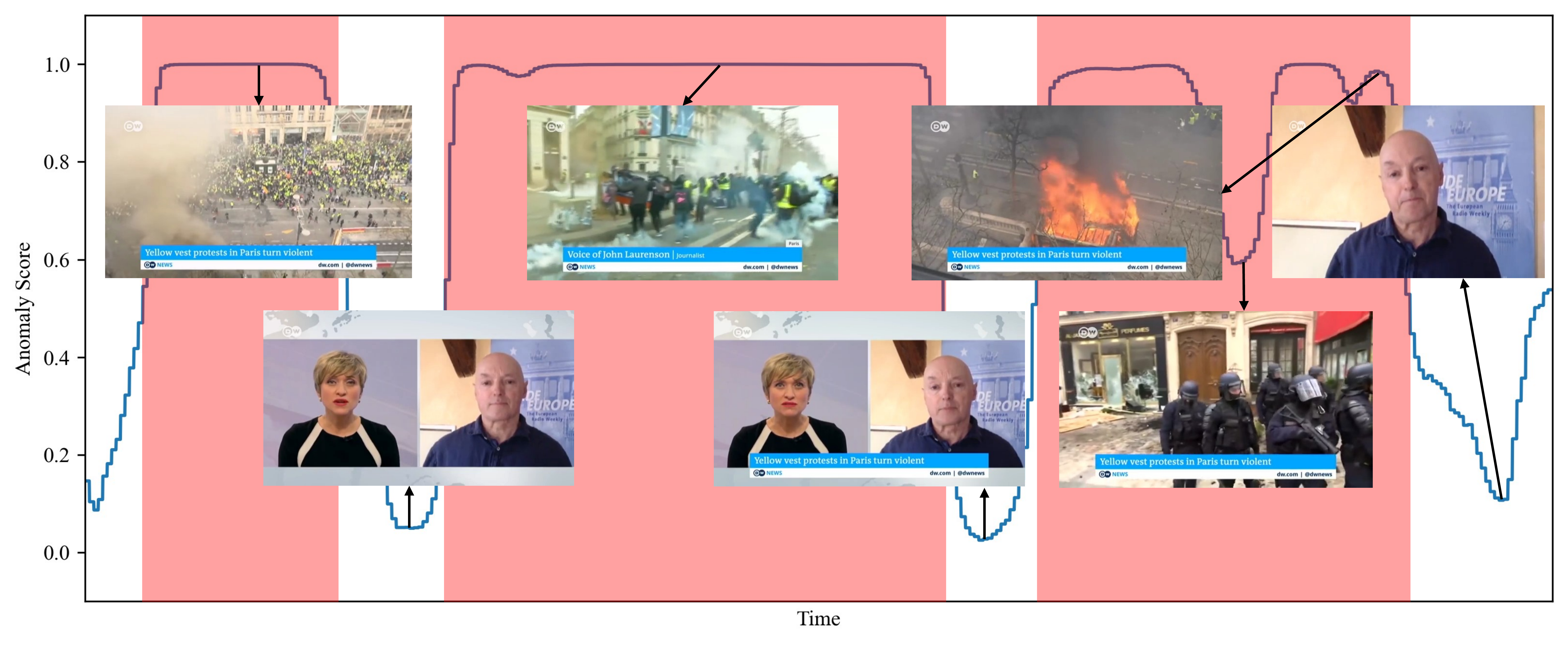}\\[-0.5ex]
		\scriptsize \textbf{(ii)} 
		\includegraphics[width=0.95\linewidth]{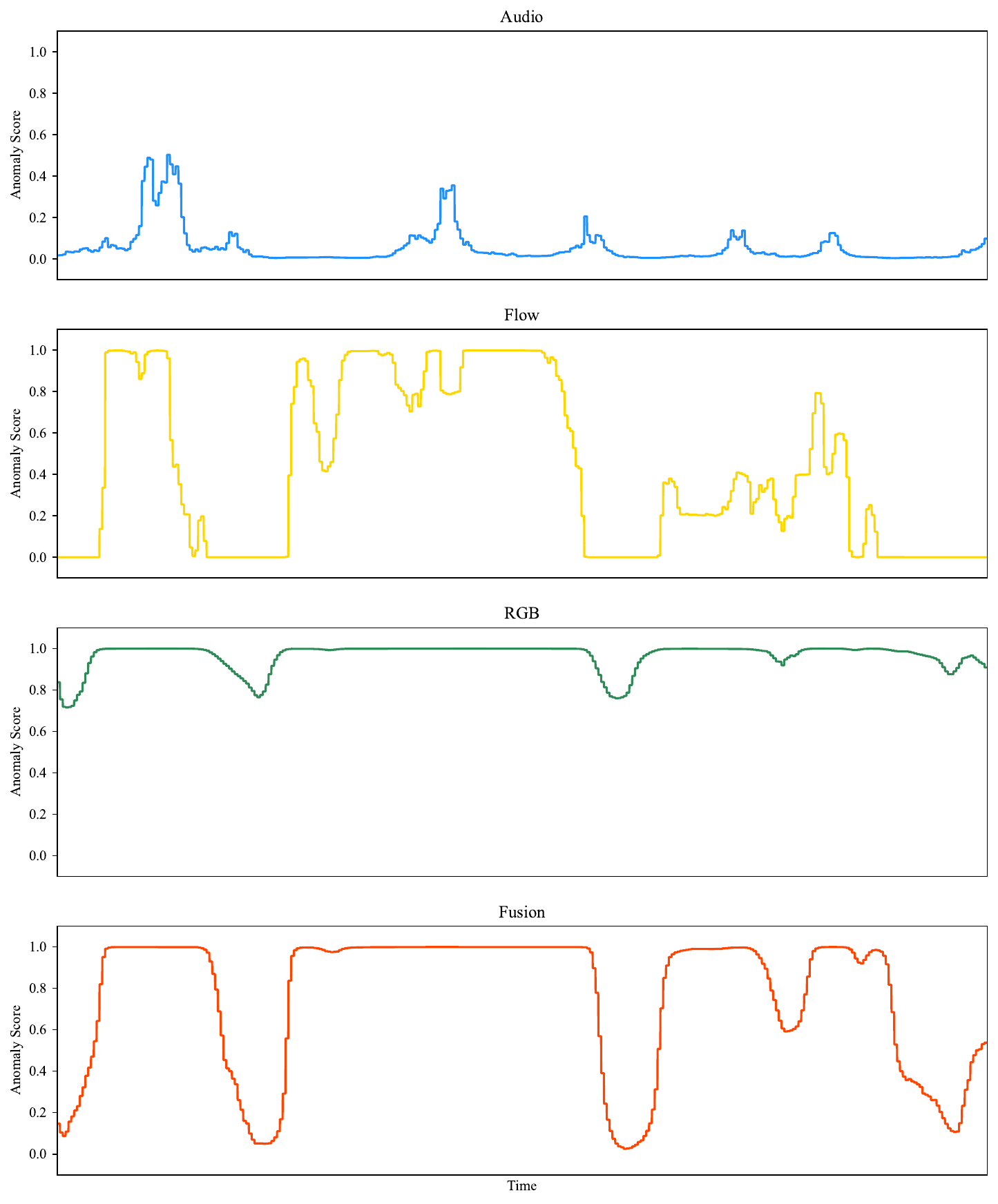}
		\subcaption{Video with riot content} 
	\end{minipage}
	\begin{minipage}{0.31\textwidth}
		\centering
		\scriptsize \textbf{(i)}
		\includegraphics[width=0.95\linewidth]{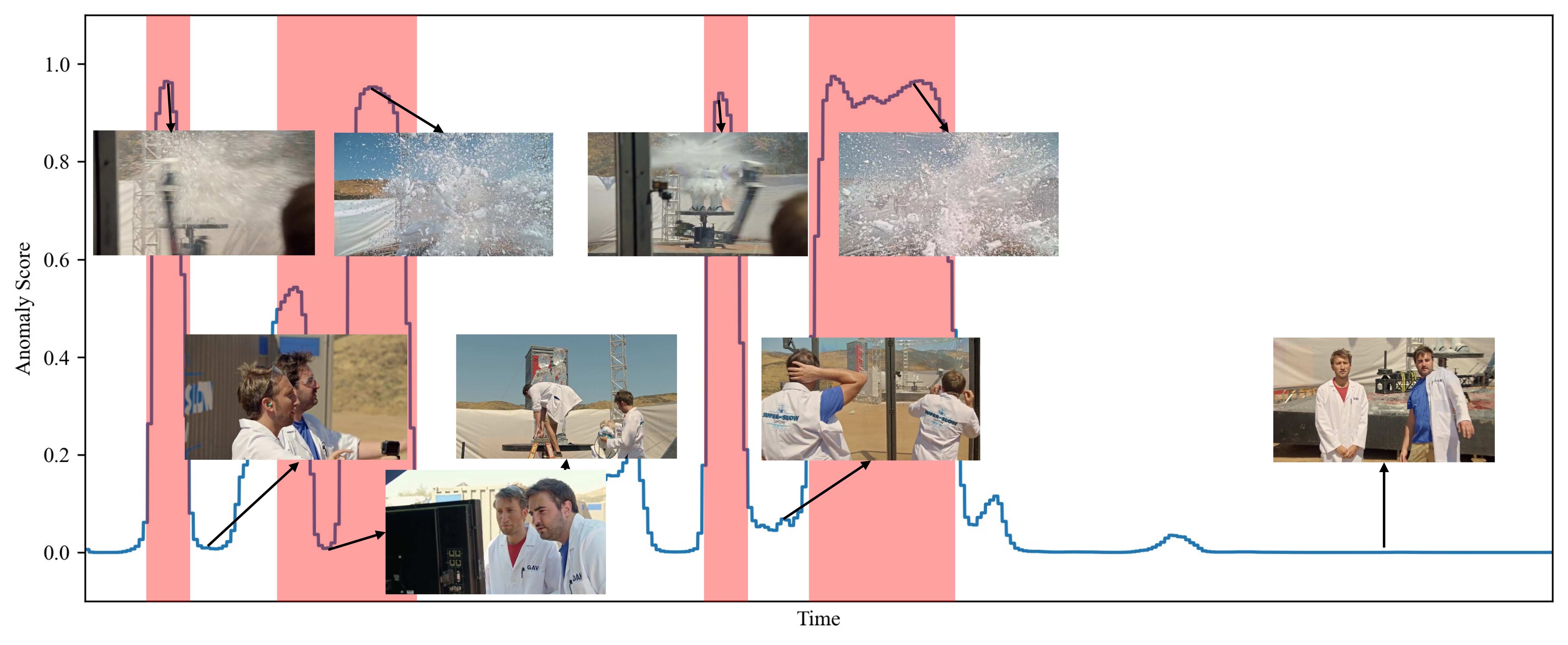}\\[-0.5ex]
		\scriptsize \textbf{(ii)}
		\includegraphics[width=0.95\linewidth]{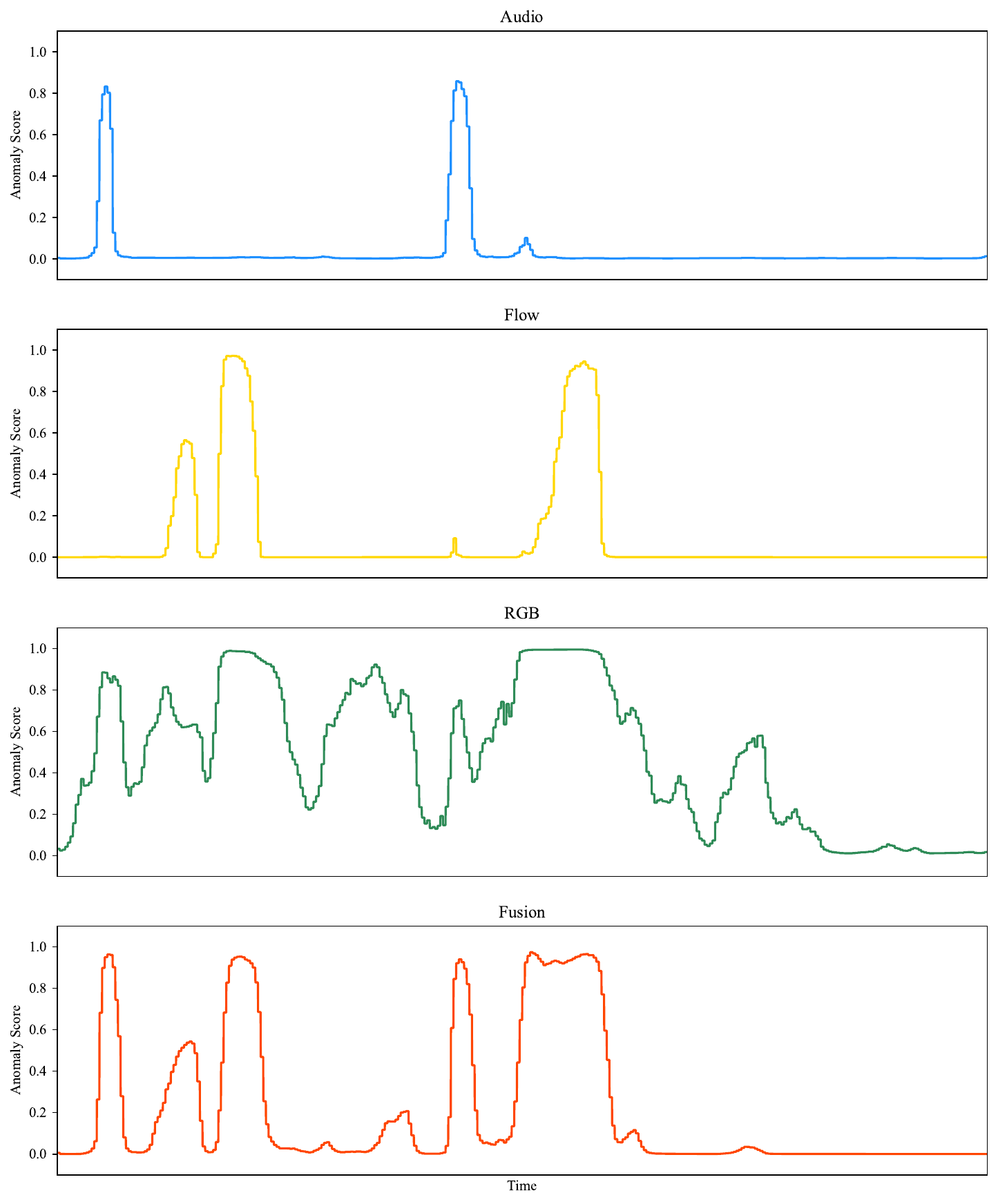}
		\subcaption{Video with explosion}
	\end{minipage}
	\begin{minipage}{0.31\textwidth}
		\centering
		\scriptsize \textbf{(i)}
		\includegraphics[width=0.95\linewidth]{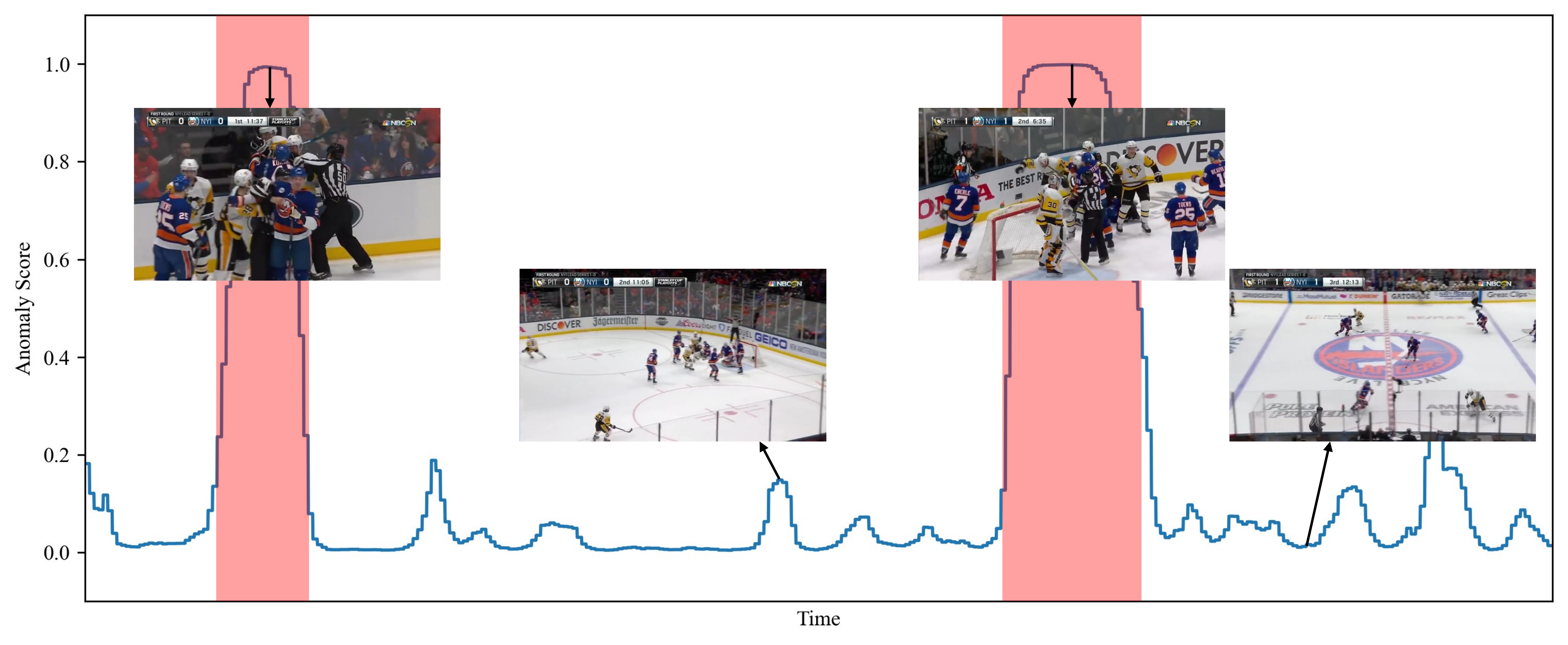}\\[-0.5ex]
		\scriptsize \textbf{(ii)}
		\includegraphics[width=0.95\linewidth]{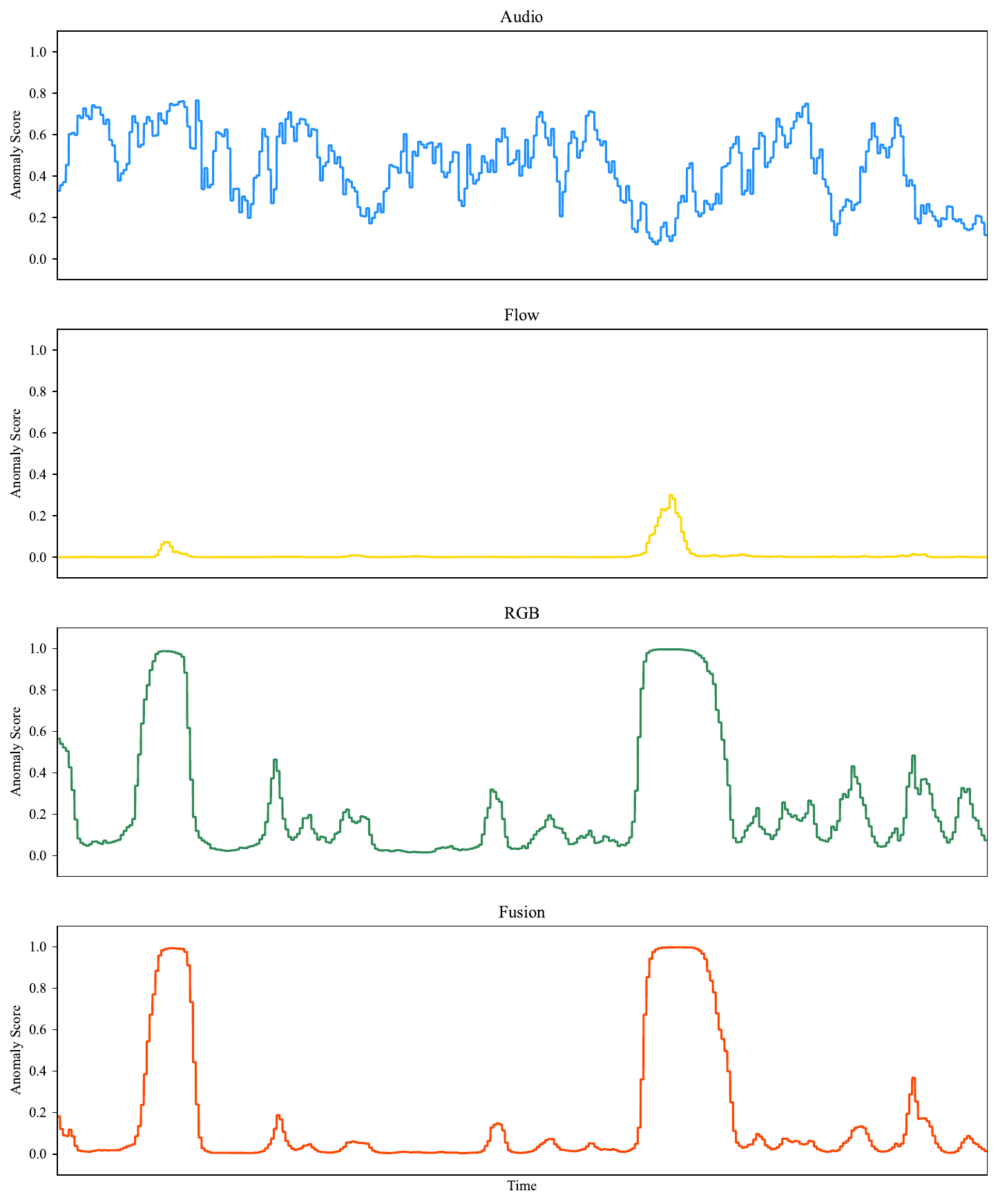}
		\subcaption{Video with fighting}
	\end{minipage}
	
	\caption{Anomaly scores of our method on XD-Violence test videos with violent content. The figure shows three test videos, each with two parts:
		\textbf{(i)} A comparison between the model's final anomaly scores output and the ground truth, with pink-highlighted regions indicating manually annotated violence events.
		\textbf{(ii)} A comparison of anomaly scores for individual modalities and the fused anomaly scores after modality fusion.}
	\label{fig:with_v}
\end{figure}

Fig. \ref{fig:with_v} presents the anomaly score curves of our method on three violent videos from the XD-Violence test set. Specifically, (a) shows riot scenes from news reports, (b) depicts explosions in TV shows, and (c) illustrates fights in ice hockey matches. Each case consists of two subplots: (i) a comparison between the model’s final anomaly scores and the ground truth labels, with pink-highlighted regions indicating the annotated violent segments; and (ii) a comparison of the anomaly score curves produced by each modality’s regression layer against the final anomaly score curve after modality fusion, where blue represents the audio modality, gold the flow modality, green the RGB modality, and red the output of the fused model.

In the riot scene video (a), the model successfully identifies the violent segments, with one exception: the third riot scene, where the anomaly score sharply decreases due to a transition to a relatively calm scene. Further analysis of subplot (ii) reveals that the audio modality is highly responsive to abnormal sounds like explosions and gunshots, while the flow modality effectively detects all riot segments but shows a weak response to the low-intensity third scene. The RGB modality, on the other hand, experiences a delay when transitioning from violent to non-violent segments. However, the fusion model effectively leverages the strengths of each modality, accurately distinguishing between violent and non-violent segments and overcoming the individual limitations of each modality.

In the explosion scene (b), the model successfully detects both regular-speed and slow-motion explosions. However, similar to video (a), the anomaly score sharply drops for the first slow-motion explosion segment due to a sudden camera switch to a non-explosive scene. Analysis of subplot (ii) shows varying performance across modalities: the audio modality struggles with slow-motion explosions due to low-frequency noise masking the acoustic features but remains highly responsive to regular-speed explosions; the flow modality, benefiting from enhanced temporal continuity in slow-motion segments, accurately captures slow-motion explosions but misses regular-speed ones; and the RGB modality experiences two false positives due to camera switching. Nevertheless, the fusion model effectively combines the strengths of each modality, accurately identifying both regular-speed and slow-motion explosions while minimizing false positives.

In the ice hockey match scene (c), the model accurately identifies the two fight segments. Analysis of subplot (ii) reveals that the audio modality's baseline anomaly score is biased high due to continuous impact sounds, resulting in many false positives. The flow modality shows minimal response during violent periods but remains stable and low during non-violent segments. The RGB modality experiences periodic noise from rapid camera switching but responds accurately to both violent segments. The fusion model, by leveraging the strengths of all modalities, effectively filters out noise, accurately identifies the violent segments, and reduces false positives.

\begin{figure}[!ht] 
	\centering
	
	\begin{tabular}{@{}c@{\hspace{0.5em}}c@{\hspace{0.5em}}c@{}}
		\subcaptionbox{Documentary clip\label{subfig:n1}}[0.3\textwidth]{
			\includegraphics[width=0.95\linewidth]{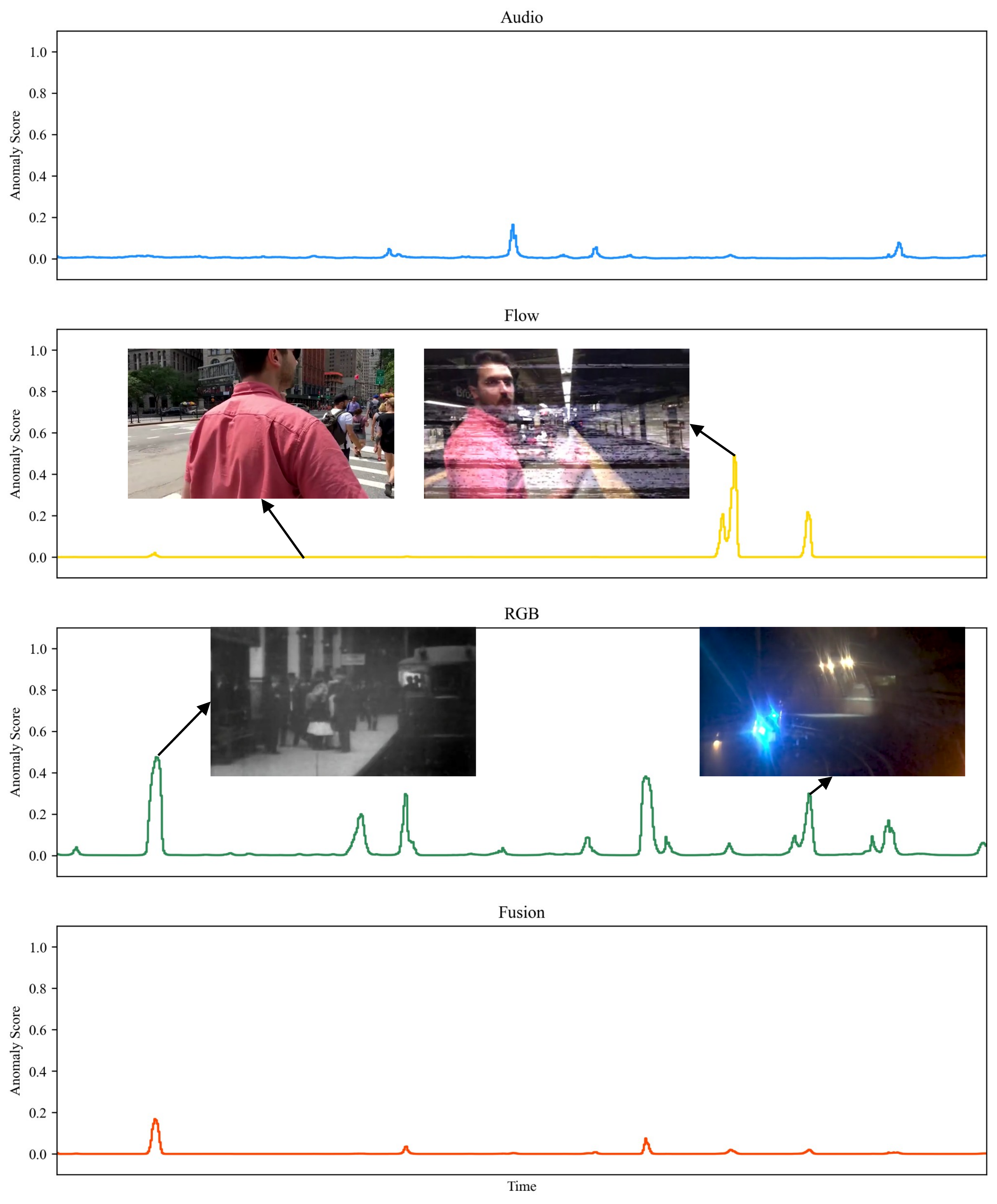}
		} & 
		\subcaptionbox{Indoor dance\label{subfig:n2}}[0.3\textwidth]{
			\includegraphics[width=0.95\linewidth]{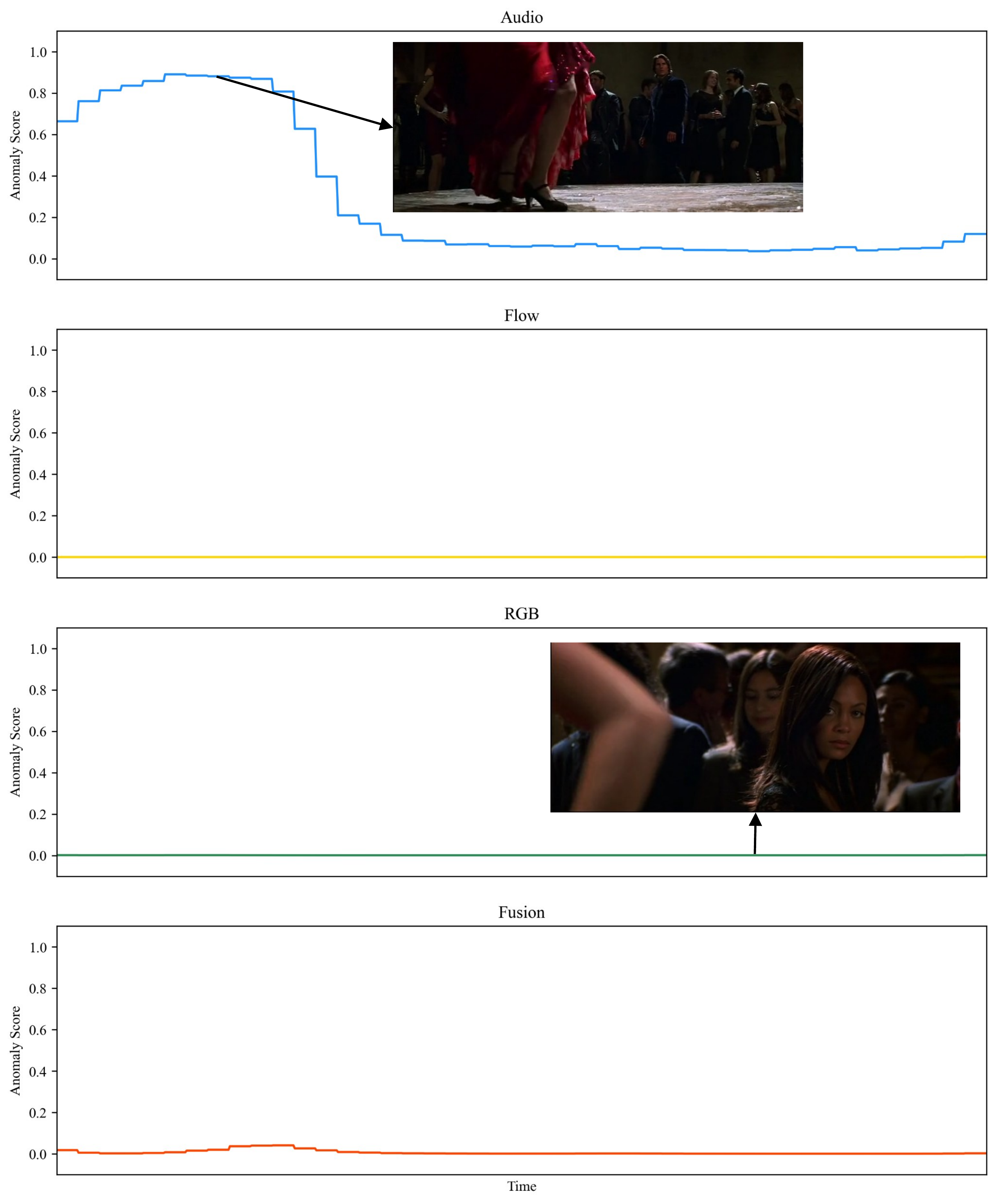}
		} & 
		\subcaptionbox{Crowd movement\label{subfig:n3}}[0.3\textwidth]{
			\includegraphics[width=0.95\linewidth]{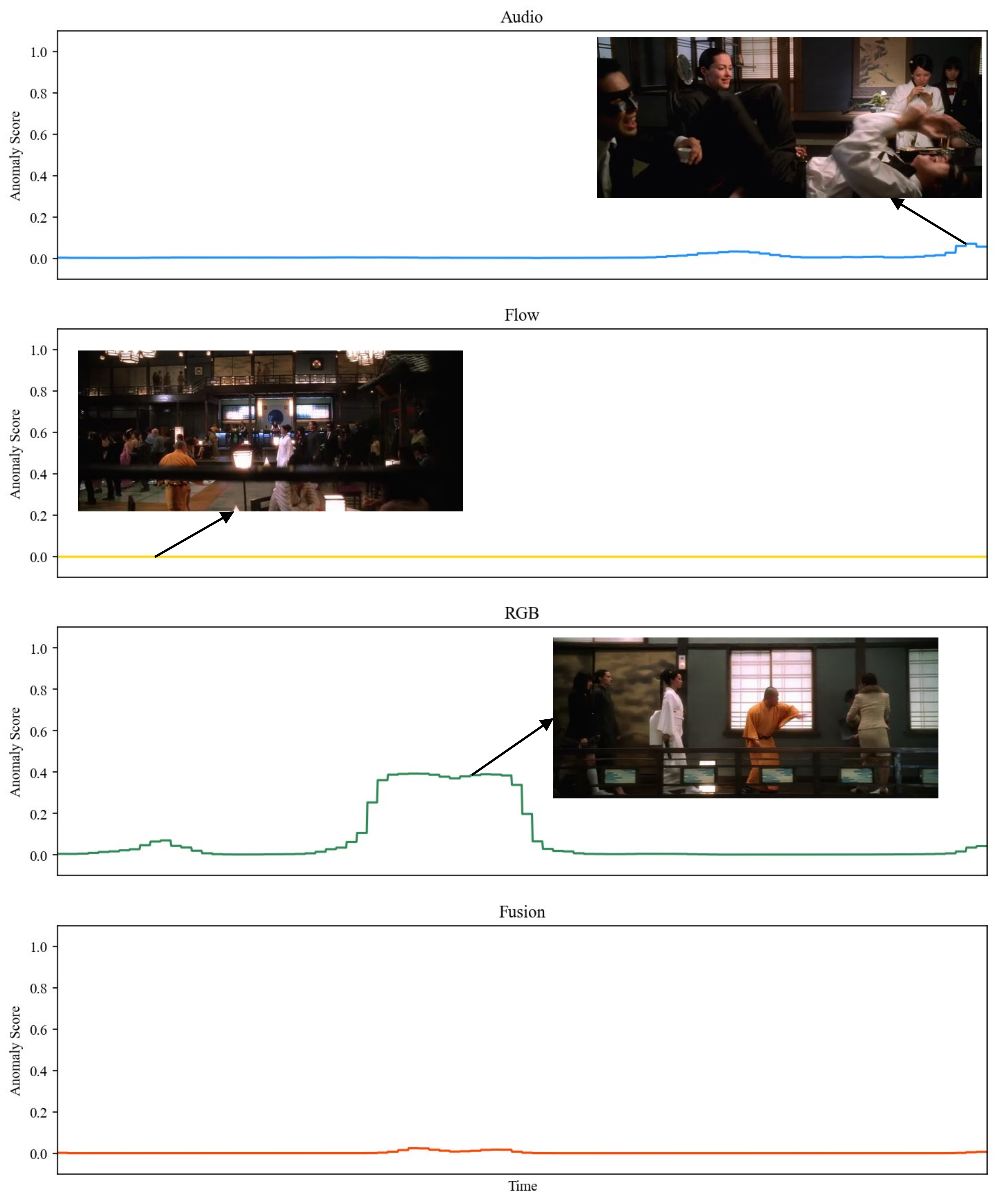}
		}
	\end{tabular}
	
	\caption{
		Anomaly score distribution on non-violent XD-Violence test scenarios. Three representative videos are shown, each containing comparative visualization of unimodal anomaly scores and fused scores after modality fusion.}
	\label{fig:with_n}
\end{figure}

Fig. \ref{fig:with_n} shows the anomaly score curves of our method on the XD-Violence non-violent test scenes. By analyzing three typical cases—documentary footage (a), indoor dance (b), and crowd movement (c)—we compare the outputs of the unimodal regression layers with the multimodal fusion results. Blue, gold, and green represent the audio, flow, and RGB modalities, respectively, while red indicates the fusion model’s outputs.

In the documentary footage (a), the audio modality remains stable. The flow modality shows a slight increase in the anomaly score during the strobe light effect used by the creator, while the RGB modality experiences a sudden spike in anomaly score due to rapid camera changes. The fusion model effectively suppresses these noises, with the largest fluctuation in anomaly scores staying under 0.2.

In the indoor dance scene (b), the audio modality’s regression layer misinterprets the frequent sounds of high heels hitting the floor, causing an abnormal increase in the anomaly score. Neither the flow nor the RGB modality responds to this. However, the fusion model successfully suppresses this misjudgment, resulting in a stable output.

In the crowd movement scene (c), both the audio and flow modalities remain stable, while the RGB modality slightly misjudges the intense motion caused by a door opening. The fusion model effectively suppresses this error, providing a more stable output.

The experimental findings from both violent and non-violent scenarios demonstrate that, after aligning the features of each modality, the fusion model effectively harnesses their strengths, thereby enhancing the model’s robustness in various complex scenarios.

\section{Conclusion}
\label{}

In this paper, we propose a novel weakly supervised multimodal violence detection method based on the principle of "Aligning First, Then Fusing." Unlike many existing methods that focus on multimodal fusion, our method prioritizes aligning modality-specific semantic features before fusion. By leveraging the inherent differences between modalities, we align the semantic features of audio, optical flow, and RGB based on their consistency in event representation, enhancing their usability. 

Specifically, the core of the alignment process involves identifying Modality-wise Feature Matching Subspaces (MFMSs) within the RGB feature space, which are most relevant to less informative modalities (e.g., audio and optical flow) in event representation. These features are then sparsely mapped into the more informative RGB space based on the MFMSs. Then, the sparse audio and optical flow features, along with the RGB features, are aligned by minimizing the distances between them. This process brings closer the features shared across modalities, strengthening the relationship between different modalities when representing the same event. This maximizes shared information, helping the fusion model better correlate complementary information, and reduces the impact of redundant data. The alignment process is dynamic and iterative, ultimately identifying the most suitable MFMSs, which results in the highest detection accuracy. 

Experimental results on the XD-Violence dataset demonstrate the effectiveness of our "Aligning First, Then Fusing" strategy, achieving a frame-level average precision (AP) of 86.07$\%$. This outperforms existing methods and underscores the advantages of our method in multimodal violence detection, providing a promising solution for weakly supervised multimodal violence detection.

\section*{Acknowledgments}
This work is supported by National Key Research and Development Program of China 
(2023YFC3321600)

\bibliographystyle{unsrt}  
\bibliography{my}

\end{document}